\documentclass[journal]{IEEEtran}

\ifCLASSOPTIONcompsoc

  \usepackage[nocompress]{cite}
\else
  \usepackage{cite}
\fi

\ifCLASSINFOpdf

\else

\fi

\usepackage{url}
\usepackage{epsfig}
\usepackage{graphicx}
\usepackage{amsmath}
\usepackage{amssymb}
\usepackage{multirow}
\usepackage{colortbl}
\usepackage{color}
\usepackage{threeparttable}
\usepackage{booktabs}
\usepackage{adjustbox}
\usepackage{subfig}
\usepackage{caption}
\usepackage[misc]{ifsym}
\usepackage{xcolor,colortbl}
\usepackage{makecell}
\usepackage{bm}
\usepackage{overpic}
\usepackage{adjustbox}
\usepackage[switch]{lineno}
\usepackage[numbers,sort&compress]{natbib}
\nolinenumbers

\newcommand*{\belowrulesepcolor}[1]{%
  \noalign{%
    \kern-\belowrulesep 
    \begingroup 
      \color{#1}%
      \hrule height\belowrulesep 
    \endgroup 
    \vspace{-0.03mm}
  }%
} 
\newcommand*{\aboverulesepcolor}[1]{%
  \noalign{%
  \vspace{-0.03mm}
    \begingroup 
      \color{#1}%
      \hrule height\aboverulesep 
    \endgroup 
    \kern-\aboverulesep 
  }%
}
\newcommand{\vspacefigtext}{\vspace{-3mm}}

\usepackage{pifont}
\newcommand{\cmark}{\ding{51}}%

\usepackage{xspace}
\makeatletter
\DeclareRobustCommand\onedot{\futurelet\@let@token\@onedot}
\def\@onedot{\ifx\@let@token.\else.\null\fi\xspace}

\usepackage[linesnumbered,lined,boxed,commentsnumbered,ruled]{algorithm2e}
\usepackage[pagebackref,breaklinks=true,colorlinks,citecolor=blue,urlcolor=blue,linkcolor=blue,bookmarks=false]{hyperref}

\begin{document}

\title{Monocular Lane Detection Based on\\ Deep Learning: A Survey}

\author{Xin He\textsuperscript{1,2}, Haiyun Guo\textsuperscript{1,2$\ast$}, Kuan Zhu\textsuperscript{1,2}, Bingke Zhu\textsuperscript{1,2}, Xu Zhao\textsuperscript{1,2}, Jianwu Fang\textsuperscript{3}, Jinqiao Wang\textsuperscript{1,2,4,5} \\
\vspace{1em}
\textsuperscript{1}School of Artificial Intelligence, University of Chinese Academy of Sciences \\
\textsuperscript{2}Foundation Model Research Center, Institute of Automation, Chinese Academy of Sciences \\
\textsuperscript{3}Institute of Artificial Intelligence and Robotics, Xi’an Jiaotong University \\
\textsuperscript{4}Peng Cheng Laboratory \qquad
\textsuperscript{5}Wuhan AI Research \\
\tt\small hexin2022@ia.ac.cn, $\{$haiyun.guo, kuan.zhu, bingke.zhu, xu.zhao, jqwang$\}$@nlpr.ia.ac.cn, \\
fangjianwu@xjtu.edu.cn \\
\IEEEcompsocitemizethanks{\IEEEcompsocthanksitem $\ast$ Corresponding Author.}
}

\IEEEtitleabstractindextext{
\begin{abstract}
Lane detection plays an important role in autonomous driving perception systems. As deep learning algorithms gain popularity, monocular lane detection methods based on them have demonstrated superior performance and emerged as a key research direction in autonomous driving perception. The core designs of these algorithmic frameworks can be summarized as follows: (1) Task paradigm, focusing on lane instance-level discrimination; (2) Lane modeling, representing lanes as a set of learnable parameters in the neural network; (3) Global context supplementation, enhancing inference on the obscure lanes; (4) Perspective effect elimination, providing accurate 3D lanes for downstream applications. From these perspectives, this paper presents a comprehensive overview of existing methods, encompassing both the increasingly mature 2D lane detection approaches and the developing 3D lane detection works. Besides, this paper compares the performance of mainstream methods on different benchmarks and investigates their inference speed under a unified setting for fair comparison. Moreover, we present some extended works on lane detection, including multi-task perception, video lane detection, online high-definition map construction, and lane topology reasoning, to offer readers a comprehensive roadmap for the evolution of lane detection. Finally, we point out some potential future research directions in this field. We exhaustively collect the papers and codes of existing works at \url{https://github.com/Core9724/Awesome-Lane-Detection} and will keep tracing the research.
\end{abstract}

\begin{IEEEkeywords}
Lane Detection, Deep Learning, Autonomous Driving
\end{IEEEkeywords}}

\maketitle

\IEEEdisplaynontitleabstractindextext

\IEEEpeerreviewmaketitle

\section{Introduction}
\label{sec:introduction}

\IEEEPARstart{L}{ane} detection seeks to obtain the semantic and positional information of each lane line from the front view (FV) image captured by the onboard monocular camera. It is an indispensable part of the perception module in autonomous driving system, providing necessary prerequisites for subsequent decision-making and planning processes. Early lane detection methods depend on manually crafted operators for feature extraction~\cite{borkar2011novel, deusch2012random, hur2013multi, jung2013efficient, tan2014novel, wu2014lane}, which have limitations in accuracy and robustness when facing the complex scenes~\cite{chiu2005lane, loose2009kalman, teng2010real, lopez2010robust, liu2010combining, zhou2010novel, jiang2009new, kim2008robust, danescu2009probabilistic}. Later on, deep learning-based methods gradually dominate this field due to their strong feature representation ability and superior performance. Nowadays, deep learning-based monocular lane detection becomes a key research topic in autonomous driving perception, attracting great attention from both academia and industry.

\begin{figure}[!t]
\centering
\includegraphics[width=\linewidth]{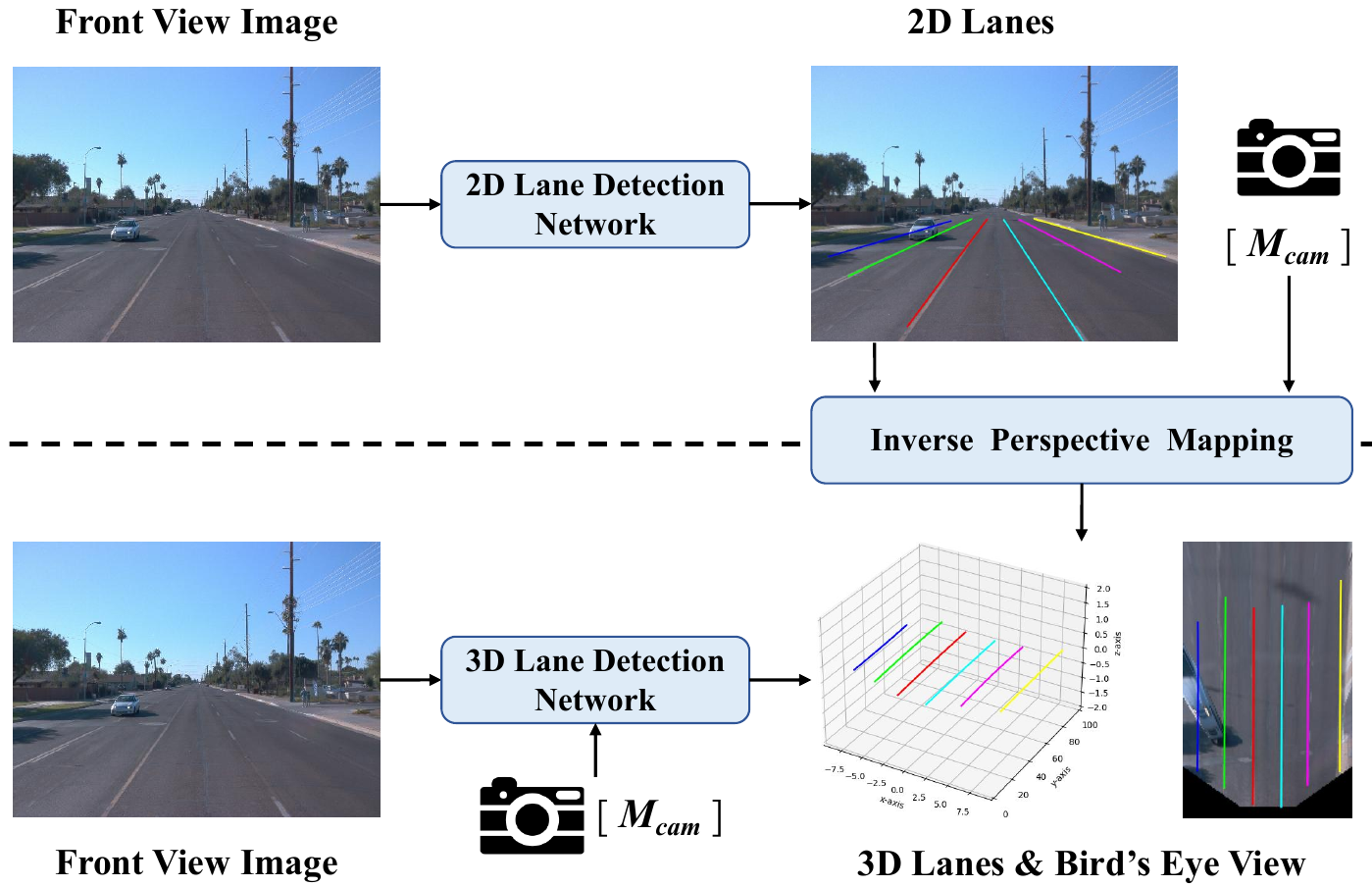}
\caption{Two different technical routes to achieve complete lane detection. Different lane instances are visualized by different colors. The 2D lane detection results need to be projected to 3D space with the help of camera parameters and inverse perspective mapping (IPM). 3D lane detection methods directly incorporate the camera parameters into the network and achieve end-to-end prediction of 3D lanes.}
\label{fig:lanedet_pipeline}
\end{figure}

\begin{figure*}[!t]
\centering
\includegraphics[width=0.95\linewidth]{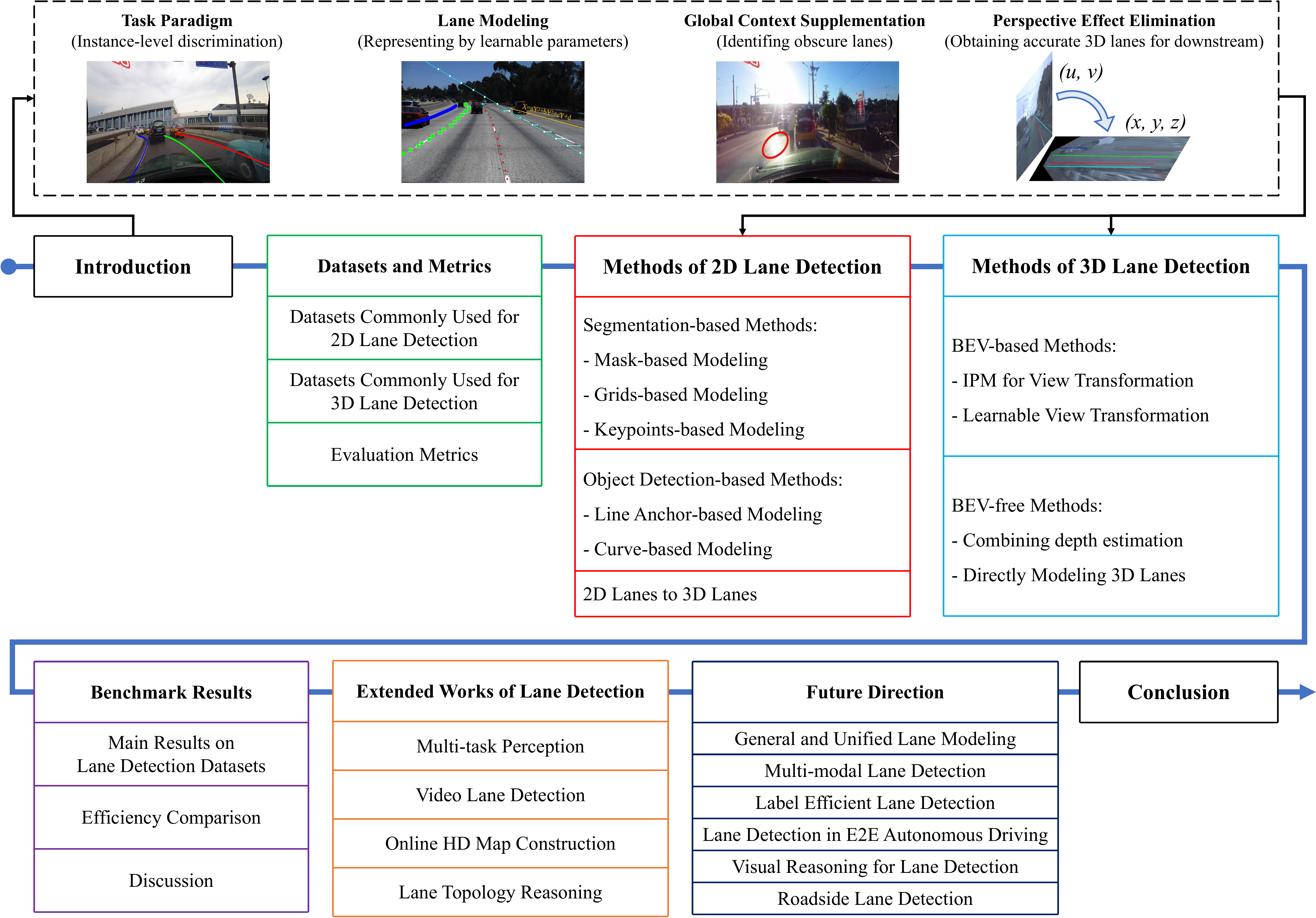}
\caption{The structure of this paper. Different colors represent specific sections. The summary of existing methods covers four core designs of the lane detection algorithms mentioned in the introduction.}
\label{fig:survey_pipline}
\end{figure*}

Existing deep learning-based monocular lane detection methods can be divided into 2D lane detection and 3D lane detection methods. As shown in Figure~\ref{fig:lanedet_pipeline}, a complete lane detection process can be described as: given a FV image, the ultimate goal is to obtain 3D lane information in the ego vehicle coordinate system, i.e., the bird's eye view (BEV) space. Because of the inherent perspective distortion in the camera imaging process, parallel lanes in the BEV plane intersect in the FV image, making it challenging to restore their true geometry. An effective solution involves building a 2D lane detection network~\cite{RESA, CondLaneNet, GANet, CLRNet, BezierLaneNet} to obtain the 2D lanes from the FV image, then the intrinsic and extrinsic parameters of the camera are combined to project these 2D lanes onto the ground through inverse perspective mapping (IPM)~\cite{IPM}, obtaining the final 3D lanes. With the rapid progress of deep learning-based methods, 2D lane detection achieve impressive results. However, IPM assumes that the ground is always flat and does not account for conditions like uphill, downhill, and rough road surfaces. It means that the IPM projection will lead the wrong 3D lane results, even if the 2D lane detection in FV is accurate. Therefore, researchers gradually shift their focus on designing 3D lane detection networks~\cite{3DLaneNet, GenLaneNet, PersFormer, Anchor3DLane, BEVLaneDet, LATR}, which directly predict 3D lanes using FV images as input.

Precise localization and real-time processing are essential for lane detection. Apart from the above, autonomous vehicles must adapt to complex road environments where lanes may be obscure due to occlusion by nearby vehicles or adverse weather conditions. Lastly, to better connect downstream tasks like planning and control, each lane instance should be distinguished and presented in a vectorized format, such as an ordered set of points or a curve equation. This is because downstream requires calculating the driving planning lines, i.e., the centerlines, based on the lanes around the vehicle. It is difficult to perform calculations without distinguishing different lane instances or lacking a vectorized representation.

Based on the complete process and the challenges of lane detection, the core design of the lane detection algorithmic frameworks can be summarized as follows: (1) \textbf{Task paradigm}, focusing on lane instance-level discrimination; (2) \textbf{Lane modeling}, representing lanes as a set of learnable parameters in the neural network; (3) \textbf{Global context supplementation}, enhancing the inference on the obscure lanes; (4) \textbf{Perspective effect elimination}, providing accurate 3D lanes for downstream applications. A comprehensive survey on lane detection should systematically explore these four key perspectives while integrating insights from both 2D and 3D lane detection methods. This dual focus is necessary to provide readers with a comprehensive understanding of advancements in lane detection technology, and help them to bridge the gap between conceptual design and practical applications.

\noindent$\bullet$
\textbf{Related surveys.}
Early reviews on monocular lane detection mainly focus on traditional methods~\cite{yenikaya2013keeping, bar2014recent}. Despite the existence of relevant summaries based on deep learning, the related surveys~\cite{tang2021lanesurvey, zhang2021lanesurvey, ma20243dlanesurvey} exhibit relatively narrow focus. On the one hand, only 2D or 3D lane detection methods are summarized. The close connection between 2D and 3D lane detection is ignored. On the other hand, the network structures or loss functions are paid too much attention in these papers, which are significant in deep learning but not the crux of addressing the lane detection challenges.

\noindent$\bullet$
\textbf{Contribution.}
This paper conducts a comprehensive investigation into the latest developments in monocular lane detection methods based on deep learning, focusing on the core designs of the lane detection algorithm frameworks. Compared to the related surveys~\cite{tang2021lanesurvey, zhang2021lanesurvey, ma20243dlanesurvey}, ours not only covers state-of-the-art 2D and 3D lane detection methods but also provides a higher-level summary. The main contributions of this survey can be summarized as follows:
\begin{enumerate}
\item We present a comprehensive survey of deep learning-based monocular lane detection methods. This is the first survey that covers both 2D lane detection and 3D lane detection.

\item This survey firstly introduces the four core designs of lane detection algorithms: task paradigm (Distinguishing different lane instances), lane modeling (representing lanes as network learnable parameters), global information supplementation (identifying obscure lanes), and eliminating perspective effects (obtaining available 3D lanes for downstream). Then we investigate the existing methods systematically from the above perspectives and summarize a general pipeline for each categorization.

\item In addition to reporting the performance of representative methods, we also reevaluate their efficiency in a unified environment. This enables readers to more easily compare different methods and select the most suitable baselines for their applications.

\item Moreover, some extended works are surveyed, including multi-task perception, video lane detection, online high-definition (HD) map construction, and lane topology reasoning. They can be regarded as an upgrade of monocular lane detection in terms of task flow. Based on these introductions, readers can receive a roadmap for the development of lane detection research focus.
\end{enumerate} 

\noindent$\bullet$
\textbf{Organization.}
The rest of the survey is organized as follows: Section~\ref{sec:Dataset_metrics} explains the datasets and evaluation metrics commonly used for Lane detection algorithms. Section~\ref{sec:methods_2d} and Section~\ref{sec:methods_3d} reviews the existing 2D and 3D lane detection methods, respectively, where we summarize existing methods from the perspective of the core designs in lane detection algorithms. Section~\ref{sec:benchmark} reports on the performance of representative methods on typical datasets and efficiency comparisons in a unified environment, and analyzes them in conjunction with the core designs of lane detection algorithms. Some expanded works of lane detection are introduced in Section~\ref{sec:extended_work}. The possible future challenges are discussed in Section~\ref{sec:future_direction} and the conclusions are provided in Section~\ref{sec:conclusion}. The sturcture of this paper is shown in Figure~\ref{fig:survey_pipline}.

\section{Datasets and Metrics}
\label{sec:Dataset_metrics}

\subsection{Datasets}
\label{sec:dataset}
Table~\ref{tab:datasets} summarizes the main statistics of prevailing lane detection benchmarks which is publicly available. Next, we provide a detailed introduction to some popular datasets.

\begin{table*}[tb]
\centering
\renewcommand{\arraystretch}{1.4}
  \setlength\tabcolsep{0.1cm}
    \caption{ 
    The overview of monocular lane detection datasets. For \textbf{Region}, ``AS'' denotes Asia, ``NA'' denotes North America, ``Sim'' denotes simulation data. For \textbf{Data Size}, ``Frames'' denotes the number of annotated and total images, ``Avg. Length'' denotes the average time duration of videos. For \textbf{Diversity}, ``Inst. Anno.'' denotes whether lanes are annotated at instance-level, ``L. Cls.'' denotes the category count (e.g. solid, dashed, etc.) of annotated lanes, and ``Max L.'' denotes the maximum number of lanes labeled in an image.
    }
\begin{center}
\resizebox{1.0\textwidth}{!}{
\begin{tabular}{l|c|c|c|c|c|c|c|c|c|c|c}
\toprule
\multirow{2}{*}{\textbf{Dataset}}  & 
\multirow{2}{*}{\textbf{Year}} &
\multirow{2}{*}{\textbf{Region}} & 
\multicolumn{2}{c|}{\textbf{Lane}} &
\multicolumn{3}{c|}{\textbf{Data Size}} &
\multicolumn{4}{c}{\textbf{Diversity}} \\
\cline{4-12}
 & & & 2D & 3D & Videos & Frames & Avg. Length & Inst. Anno. & L. Cls. & Max L. & Resolution \\
\midrule
Caltech~\cite{Caltech} & 2012 & NA & \cmark & - & 4 & 1224/1224 & - & \cmark & - & 4 & 640$\times$480 \\
VPGNet~\cite{VPGNet} & 2017 & AS & \cmark & - & - & 20K/20K & - & - & 7 & - & 640$\times$480 \\
Tusimple~\cite{Tusimple} & 2017 & NA & \cmark & - & 6.4K & 6.4K/128K & 1s & \cmark & - & 5 & 1280$\times$720 \\
CULane~\cite{SCNN} & 2018 & AS & \cmark & - & - & 133K/133K & - & \cmark & - & 4 & 1640$\times$590 \\
ApolloScape~\cite{Apolloscape} & 2018 & Sim & \cmark & - & 235 & 115K/115K & 16s & - & 13 & - & 3384$\times$2710 \\
BDD100K~\cite{BDD100K} & 2018 & NA & \cmark & - & 100K & 100K/120M & 40s & - & 11 & - & 1280$\times$720 \\
LLAMAS~\cite{LLAMAS} & 2019 & NA & \cmark & \cmark & 14 & 79K/100K & - & \cmark & - & 4 & 1276$\times$717 \\
CurveLanes~\cite{CurveLanes} & 2020 & AS & \cmark & - & - & 150K/150K & - & \cmark & - & 9 & 2560$\times$1440 \\
Apollo 3DLane~\cite{GenLaneNet} & 2020 & Sim & \cmark & \cmark & - & 10K/10K & - & \cmark & - & 6 & 1920$\times$1080 \\    
VIL-100~\cite{VIL100} & 2021 & AS & \cmark & - & 100 & 10K/10K & 10s & \cmark & 10 & 6 & 640$\times$368$\sim$1920$\times$1080 \\
Comma2k19-LD~\cite{Comma2k19LD} & 2022 & NA & \cmark & - & 100 & 2K/2K & 1s & \cmark & - & - & 1164$\times$874 \\
SDLane~\cite{Eigenlanes} & 2022 & AS & \cmark & - & - & 45K/45K & - & \cmark & - & 4 & 1920$\times$1208 \\
ONCE-3DLanes~\cite{ONCE3DLanes} & 2022 & AS & \cmark & \cmark & - & 211K/211K & - & \cmark & - & - & 1920$\times$1020 \\
OpenLane~\cite{PersFormer} & 2022 & NA & \cmark & \cmark & 1K & 200K/200K & 20s & \cmark & 14 & 24 & 1920$\times$1280 \\
CarLane~\cite{CarLane} & 2022 & Sim$\&$NA & \cmark & - & - & 118K/163K & - & \cmark & - & 4 & 1280$\times$720 \\
OpenLane-V~\cite{RVLD} & 2023 & NA & \cmark & - & 590 & 90K/90K & 20s & \cmark & - & 4 & 1920$\times$1280 \\
\bottomrule
\end{tabular}}
\end{center}
\label{tab:datasets}
\vspace{-0.4cm}
\end{table*}

\subsubsection{Datasets Commonly Used for 2D Lane Detection}
\noindent$\bullet$
\textbf{Tusimple.} 
The TuSimple~\cite{Tusimple} dataset is collected with stable lighting conditions in highways, including different levels of occlusion, different types of lanes, and different road conditions. It consists of 6,408 images, which are split into 3,268 training, 358 validation, and 2,782 test images. For each image, lanes are annotated by the 2D coordinates of sampling points with a uniform height interval of 10 pixels. Each annotated image has a size of 1280$\times$720 pixels.

\smallskip
\noindent$\bullet$
\textbf{CULane.} 
CULane~\cite{SCNN} is a large-scale 2D lane detection dataset with 88,880 training images and 34,680 testing images. In addition to different weather conditions and light levels, there are eight challenging lane detection scenarios, such as traffic congestion, shadow occlusion, missing lanes, and lane curves. All the images have 1640$\times$590 pixels.

\smallskip
\noindent$\bullet$
\textbf{LLAMAS.} 
The annotations of LLAMAS~\cite{LLAMAS} are automatically generated from HD maps.  This dataset contains over 100k images from about 350 km of recorded drives. In contrast to other datasets, LLAMAS presents a small and variable number of pixels marking each lane, reflecting real-world conditions more accurately. The resolution of all images is 1280$\times$717 pixels.

\smallskip
\noindent$\bullet$
\textbf{CurveLanes.} 
CurveLanes~\cite{CurveLanes} contains 100K, 20K, and 30K images for training, validation, and testing, respectively. It features an abundance of curved lanes and difficult scenarios such as S-curves, fork lines, nighttime conditions, and multi-lane configurations. In comparison to existing datasets like the first three, each image within CurveLanes encompasses a greater number of lanes and has a higher resolution.

\subsubsection{Datasets Commonly Used for 3D Lane Detection}
\noindent$\bullet$
\textbf{Apollo 3DLane.} 
The Apollo 3DLane dataset~\cite{GenLaneNet} is generated using the game engine, including 10,500 discrete frames of monocular RGB images and their corresponding 3D lanes ground truth, which is split into three scenes: balanced, rarely observed, and visual variation scenes. Each scene contains independent training sets and test sets.

\smallskip
\noindent$\bullet$
\textbf{ONCE-3DLanes.} 
ONCE-3DLanes~\cite{ONCE3DLanes} is a large-scale real-world 3D lane detection dataset, which is constructed based on the ONCE dataset~\cite{ONCE}. It contains 211K images comprising diverse weather conditions (sunny, cloudy, rainy) and varied geographical locations (urban centers, suburban areas, highways, bridges, and tunnels).  Only intrinsics of the camera are provided in ONCE-3DLanes.

\smallskip
\noindent$\bullet$
\textbf{OpenLane.} 
OpenLane~\cite{PersFormer} is another large-scale but more comprehensive benchmark for real-world 3D lane detection based on Waymo Open Dataset~\cite{Waymo}. The dataset includes 200K images captured in a variety of weather, terrain, and brightness conditions. In OpenLane, the lane annotation not only contains the 3D position of a lane but also several attributes and tracking id. The intrinsics and extrinsics of the camera are provided for each frame, and category and scene labels (e.g., weather and location) are also provided, providing a realistic and diverse set of challenges for 3D lane detection algorithms.

\subsection{Evaluation Metrics}
\label{sec:metric}
Despite that different datasets may utilize ununified evaluation metrics, below we mainly introduce the common evaluation metrics adopted by all datasets. More evaluation metrics can be found in the Section~\ref{sec:more_metric} of Appendix.

F1 score serves as the primary metric, taking into account both accuracy and recall. The calculation of recall and correct rate is closely related to the determination of true positive (TP). Different datasets determine TP in different ways. The F1 score is calculated as follows:

\begin{equation}\label{F1score}
F_1 = \frac{2 \times Precision \times Recall}{Precision + Recall}. 
\end{equation}

\begin{equation}\label{precision&recall}
Precision = \frac{TP}{TP + FP}; \quad Recall = \frac{TP}{TP + FN}.
\end{equation}

Tusimple~\cite{Tusimple} and CULane~\cite{SCNN}, which are the representive 2D lane detection benchmarks, adopts two different ways to determine TP. Tusimple~\cite{Tusimple} focuses on point-by-point evaluation. The predicted point is considered correct if the horizontal distance from the true value point is less than 20 pixels when the longitudinal coordinates are the same. Furthermore, the line prediction is viewed as TP when it contains no less than 85\% of the true value points. In contrast, CULane~\cite{SCNN} emphasizes line-by-line evaluation, treating each lane as a mask of several pixels wide, and calculating the intersection (IoU) between the predicted lane and the annotated lane. The prediction with IoU larger than 75\% is viewed as TP.

For 3D lane detection, there are also two main ways to determine the TP, represented by the evaluation methods in OpenLane~\cite{PersFormer} and ONCE-3DLanes~\cite{ONCE3DLanes}, respectively. OpenLane~\cite{PersFormer} follows the evaluation metric designed by~\cite{GenLaneNet}. The matching between prediction and ground truth is built upon edit distance, where one predicted lane is considered to be a TP only if 75\% of its covered y-positions have a point-wise distance less than the max-allowed distance (1.5m). The ONCE-3DLanes~\cite{ONCE3DLanes} dataset employs a two-stage evaluation metric for lane detection. First, the IoU method from CULane~\cite{SCNN} is utilized on the z-x plane (i.e., top view) to assess the alignment between the prediction and ground truth. Second, if the IoU exceeds a predefined threshold, the curve matching error in camera coordinates is computed using unilateral chamfer distance. If this unilateral chamfer distance falls below the specified threshold, the prediction is classified as TP.

\section{Methods of 2D Lane Detection}
\label{sec:methods_2d}
This section reviews the existing 2D lane detection methods. We first explain the ground for classifying existing methods in Section~\ref{sec:taxonomy_2d}, and then discuss the existing methods accordingly in Section~\ref{sec:segmentation-based methods} and Section~\ref{sec:object detection-based methods}. Lastly, the process of converting 2D lanes to 3D lanes using IPM and its deficiencies are reviewed in Section~\ref{sec:IPM}.

\subsection{Classification Framework}
\label{sec:taxonomy_2d}
Previous summaries~\cite{tang2021lanesurvey, zhang2021lanesurvey} primarily focus on the design of network structures and loss functions. It is noteworthy that the instance-level discrimination and vectorized result representation, which are necessary prerequisites for guiding downstream applications, are overlooked. By contrast, our classification of 2D lane detection methods is primarily according to the above two aspects. 

\begin{figure*}[t]
\centering
\includegraphics[width=1 \textwidth]{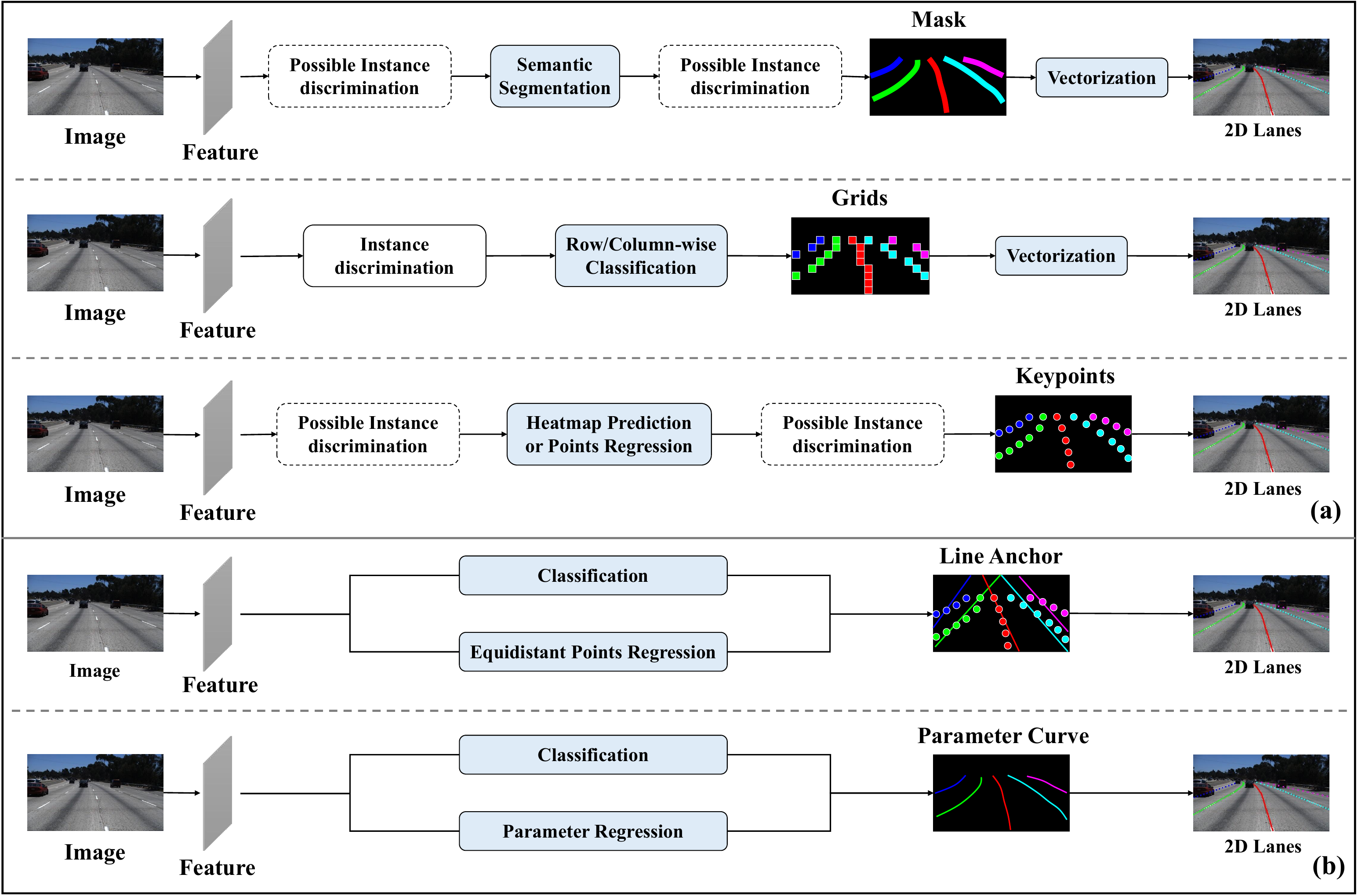}
\caption{
The general pipelines for 2D lane detection: (a) Segmentation-based methods, which leverage mask, grids, or keypoints to model a lane, perform instance discrimination and lane localization sequentially. Each method only performs instance discrimination once. (b) Object detection-based methods, which leverage line anchor or parameter curve to model a lane, complete the instance discrimination and lane localization in parallel.
}
\label{fig:methods_2d}
\vspace{-0.4cm}
\end{figure*}

\begin{figure}[tbp]
\centering
\includegraphics[width=0.48 \textwidth]{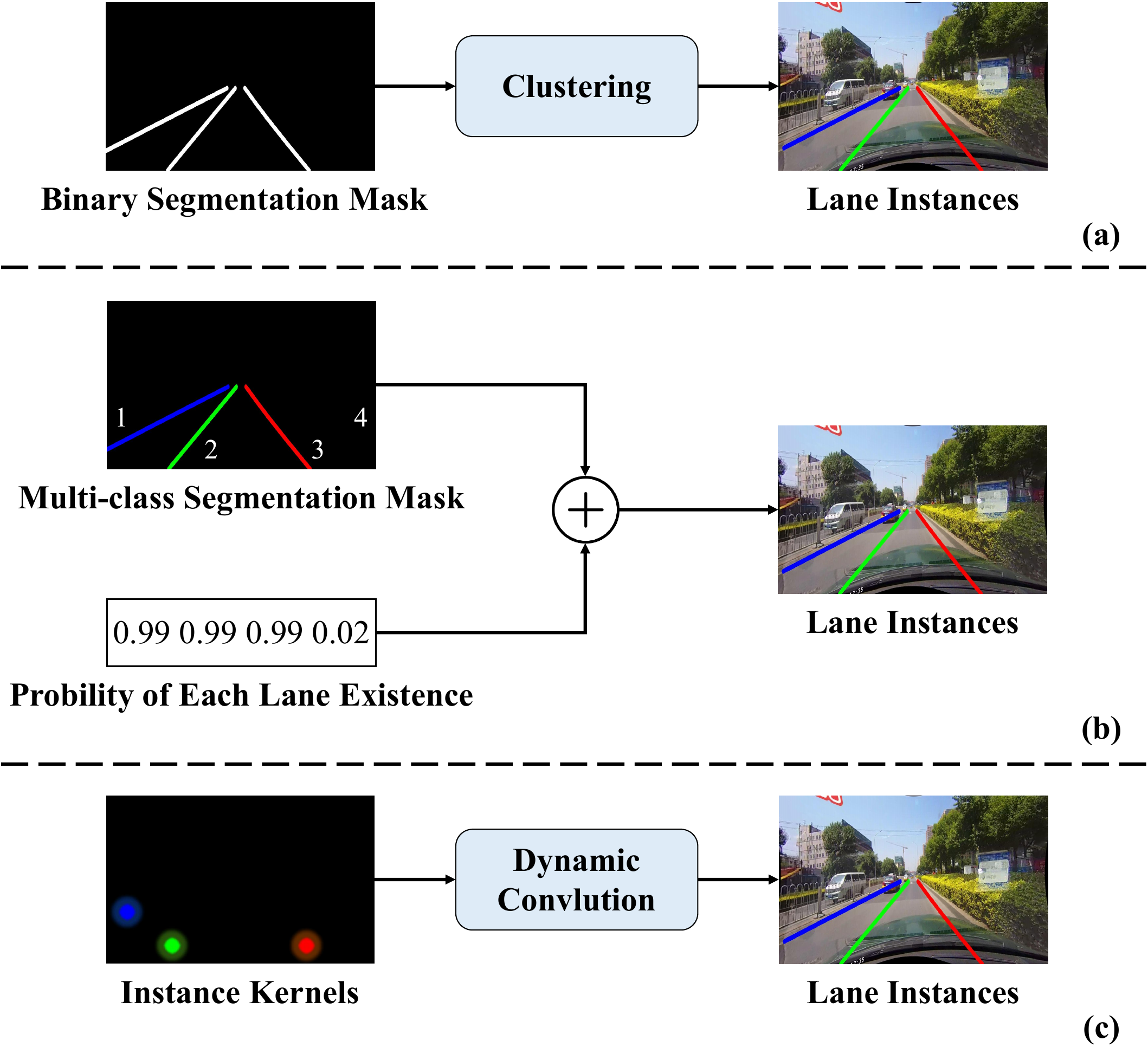}
\caption{
The ways for discriminating different lane instances in segmentation-based methods: (a) Bottom-up faction represented by~\cite{LaneNet, LaneAF, FOLOLane, GANet}. (b) Top-down faction represented by~\cite{SCNN, SAD, RESA, UFLD, GANet}, which predefines the maximum number of lanes and treats each lane as a category. (c) Top-down faction represented by~\cite{CondLaneNet, CondLSTR}, which predicts dynamic kernels to generate instances. Among them, grids-based modeling is only applicable to the top-down approach.
}
\label{fig:instance_discrimination}
\vspace{-0.4cm}
\end{figure}

As shown in Figure~\ref{fig:methods_2d}, first, for lane instance-level discrimination, 2D lane detection methods can be divided into two types of paradigms based on the number of stages required to complete the task: (a) \textbf{Segmentation-based methods} (two-stage), which complete the lane localization and instance discrimination in a certain order. Figure~\ref{fig:instance_discrimination} summarizes the general pipelines for instance-level discrimination in such methods. (b) \textbf{Object detection-based methods} (one-stage), which perform instance discrimination and localization concurrently. This advantage arises from the general pipeline of object detection algorithms, which execute both classification and regression tasks on a set of candidate proposals in parallel.

\begin{figure}[tbp]
\centering
\includegraphics[width=0.48 \textwidth, trim={2cm 0.5cm 2cm 0cm}, clip]{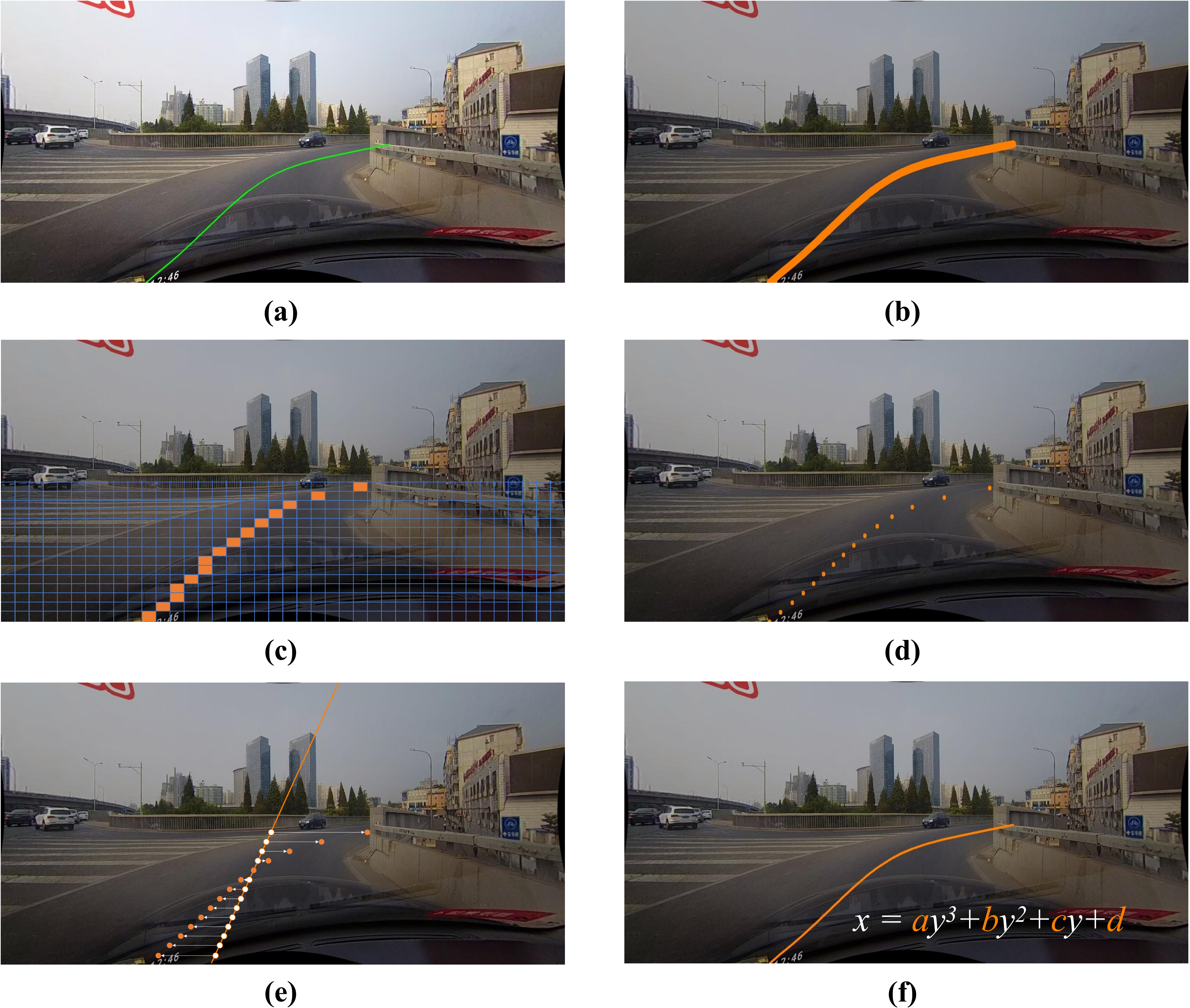}
\caption{
More details of different lane modeling. (a) A lane in FV image. (b) Mask-based modeling, which performs semantic segmentation. (c) Grids-based modeling, which executes the row/column-wise classification. (d) Keypoints-based modeling, which detects discrete points on the lane. (e) Line Anchor-based modeling, which learns the horizontal offset of equidistant points. (f) Curve-based modeling, which predicts the curve parameters.
}
\label{fig:lane_modeling}
\vspace{-0.4cm}
\end{figure}

Second, vectorized result representation requires algorithms to consider how to model lanes as a set of values for neural network learning, i.e., lane modeling. In terms of lane modeling, the segmentation-based methods can be further divided into \textbf{mask-based modeling}, \textbf{grids-based modeling}, and \textbf{keypoints-based modeling}. For object detection-based methods, adopting a bounding box to model a narrow and long lane is often not reasonable. This is because the bounding boxes generated by object detection methods may be mutually occluded, and a bounding box may contain multiple lane instances. To align with the general object detection paradigm, these methods design unique "bounding boxes" to model lanes, including \textbf{line anchor-based modeling} and \textbf{curve-based modeling}. The details of each lane modeling are described in Figure~\ref{fig:lane_modeling}.

Furthermore, most existing 2D lane detection datasets provide complete annotations for lanes, even the lanes are severely occluded by vehicles or affected by extreme weather conditions. To better identify such obscure lanes, many algorithms meticulously design special structures within their networks, thus the measures they use are also described in this section. We elaborately compare the representative 2D lane detection works according to the above classification criteria in Table~\ref{tab:2d_lane_detection_methods}.

\subsection{Segmentation-based Methods}
\label{sec:segmentation-based methods}
\subsubsection{Mask-based modeling}
\label{sec:mask}

Given an image \(I\in R^{H \times W \times 3}\), the ultimate goal of the network is to predict a set of masks of the same size as the input image. In the early stages, fully convolutional segmentation networks represented by~\cite{FCN} are used to segment lanes~\cite{zang2018traffic}. The encoder extracts high-level semantic information into feature maps, and then the decoder upsamples these feature maps to their original size for pixel-wise prediction. However, even with more powerful general segmentation networks~\cite{Deeplab,  ENet, ERFNet, BiseNet, EDANet}, the lane segmentation performance remains unsatisfactory. This is mainly because these networks do not account for the elongated nature of lanes. Moreover, when lanes are occluded by factors like vehicle or lighting, relying solely on annotations of complete lanes for supervision is not an effective solution. Traditional encoders often fail to capture these subtle features. Consequently, numerous studies introduce specialized structures before pixel-wise prediction to enhance feature representation.

VPGNet~\cite{VPGNet} predicts the disappearance points of lanes as the global geometric background to improve the performance of lane detection. SCNN~\cite{SCNN} develops a novel convolutional layer for specially shaped objects, such as lanes and utility poles, allowing information to pass between image layers, which is similar to recurrent neural networks. However, there is still some room for improvement in computational speed. SAD~\cite{SAD} employs a self-attention distillation mechanism that contextually aggregates high-level and low-level attention to obtain finer lane features. IntRA-KD~\cite{IntRA-KD} uses a teacher-student distillation mechanism to represent lane structure knowledge as an interregional affinity map, capturing the similarity of lane feature distribution across different scene regions. EL-GAN~\cite{EL-GAN} uses the generative adversarial network (GAN) to obtain more realistic and structurally rich lane segmentation results. Then, Zhang et al.~\cite{Ripple-GAN} select a GAN with better performance~\cite{WGAN-GP} and modify its structure to extract subtle lane features. Xu et al.~\cite{SALMNet} design a channel attention module that enhances lane features and suppresses background noise, and propose a pyramid deformation convolution module to obtain more structural information of lanes. RESA~\cite{RESA} further proposes a recurrent aggregator on top of SCNN~\cite{SCNN} that fully exploits the lane shape prior to enable the network to aggregate global features for improved performance and efficiency. PriorLane~\cite{PriorLane} obtains more comprehensive features based on a Mixed Transformer~\cite{SegFormer} and improves network performance by fusing image features with low-cost local prior knowledge, enhancing lane segmentation.

While semantic segmentation provides semantic categories at the pixel level, it is insufficient for distinguishing different instances within the same category. An intuitive approach is to apply top-down instance segmentation frameworks, such as Mask R-CNN~\cite{MaskRCNN} or YOLACT~\cite{YOLACT}, to achieve instance-level discrimination and segmentation of lanes. However, the bounding boxes generated by object detection methods may contain multiple lane instances, which complicates distinguishing them in the subsequent semantic segmentation process.t
 
SCNN~\cite{SCNN} proposes a top-down process that is different from the above. Specifically, each lane is treated as a separate category so that multi-category semantic segmentation is performed. Meanwhile, a parallel classification branch is incorporated to predict the existence of lanes at each position. Finally, the classification and segmentation results are combined to obtain the final lanes. The subsequent works~\cite{EL-GAN, SAD, RESA, PriorLane} follow this way. This manner facilitates instance differentiation but introduces certain limitations: it requires defining a maximum number of lanes in advance to determine the number of possible instances. Additionally, the correspondence between lanes and classes is established by annotations. When vehicles switch between lanes, this predefined labeling may lead to ambiguity.

To solve the above problem, some studies adopt a bottom-up approach for instance segmentation, i.e., cluster the binary segmentation results of lanes/backgrounds. VPGNet~\cite{VPGNet} clusters lanes using a modified density-based clustering method. LaneNet~\cite{LaneNet} utilizes instance embedding to cluster the results of semantic segmentation, achieving lane instance segmentation. This method offers high clustering accuracy but is time-consuming, which limits its applicability for real-time processing. FastDraw~\cite{FastDraw} constructs a learnable decoder that not only segments lanes but also identifies pixels belonging to the same lane. To address the inefficiency of pixel-embedding clustering, LaneAF~\cite{LaneAF} introduces an affinity vector field to associate pixels belonging to the same lane. Although these methods are more flexible, the algorithm execution efficiency remains suboptimal due to the high complexity of bottom-up clustering and the low efficiency of mask-based modeling in classifying all pixels.

The lane masks obtained from the segmentation network usually contain a large number of irrelevant areas. In order to be used for ego-vehicle motion prediction and planning, it is necessary to further denoise the mask to obtain vectorized results. Usually, for each lane mask, the highest response is sampled sequentially at equidistant heights, and then curve fitting is performed.

\subsubsection{Grids-based modeling}
\label{sec:grids}
To address the inefficiency of pixel-wise prediction in semantic segmentation, UFLD~\cite{UFLD} proposes a grids-based modeling approach. It divides the image into $h$ rows and $w$ columns, creating \( h \times w \) grids with equal spacing along the height and width. Lane detection is then described as a row-wise prediction process. For each lane instance, a grid that is most likely to belong to it is predicted in each row. In this way, the original pixel-by-pixel classification requires a time complexity of \(O(H \times W \times C)\), while this method reduces the complexity to \(O(h \times w \times C)\), where C is the number of classes. It is clear that \(h \ll H\) and \(w \ll W\). Therefore the ultra-fast inference is enabled. To supplement global context, the network selects a large fully connected (FC) layer to output the classification probabilities for each grid, thereby increasing the receptive field.





For lane instance discrimination, the premise is the existence of known instances, which means it cannot be performed in the bottom-up manner. UFLD~\cite{UFLD} follows SCNN~\cite{SCNN} by treating each instance as a category, which is not robust. To solve the instance discrimination problem, and inspired by instance segmentation methods like CondInst~\cite{CondInst} and SOLOv2~\cite{SOLOv2}, CondLaneNet~\cite{CondLaneNet} learns the probabilistic heatmaps of the lane starting points to obtain lane instances and generates dynamic kernels based on the features of the starting points. Then, conditional convolution~\cite{CondConv} is applied to the kernel and the entire feature map for row-wise classification. Additionally, a recurrent instance module based on LSTM~\cite{LSTM} is proposed to address dense lines and forked line scenarios.

The strategy of row-wise classification leverages the vertical and slender nature of lanes. Unfortunately, it is not well-suited for some curved or near-horizontal lanes, where several meshes in a row may correspond to the same lane. For this reason, UFLD-V2~\cite{UFLDv2} extends row-wise classification to row/column-wise classification to address the issue that row-wise classification cannot handle horizontal lanes. However, it still employs a multi-classification strategy~\cite{SCNN} for lane instance discrimination, which results in an overly simplistic choice between row and column classification, thereby limiting its generalizability in real-world scenarios. CANet~\cite{CANet} further optimizes this approach. It employs U-shaped guidelines to constrain lane instance kernel generation based on CondLaneNet~\cite{CondLaneNet}. An adaptive decoder is designed, which dynamically chooses between row-by-row or column-by-column classification for each instance.

Since the networks output the classification probability of each row/column grid, rather than the vectorized format, the post-processing is also required. Specifically, each point’s coordinate is calculated as the expectation of locations (grids from the same row/column), i.e., a weighted average by probability. Compared to the post-processing of lane masks obtained from semantic segmentation, it is easier to implement.

\begin{table*}[t]
\scriptsize
\renewcommand\arraystretch{1.2}
\centering
\caption{The summary of representative 2D lane detection methods. For \textbf{Task Paradigm}, "Seg" means segmentation-based methods, "ODet" means object detection-based methods. For segmentation-based methods, the manner of instance discrimination is additionally indicated instead of using "\cmark" to mark. "$\uparrow$" and "$\downarrow$" represents the instance discrimination via bottom-up and top-down approach in segmentation-based methods, respectively. "Max. L." represents predefining the maximum number of lanes, and "Dy. K." represents predicting dynamic kernels.}
\begin{center}
    \resizebox{1.0\textwidth}{!}{
    \begin{tabular}{l|c|c|c|c|c} 
    \toprule
    \multirow{2}{*}{\textbf{Methods}} & \multirow{2}{*}{\textbf{Venue}} & \multicolumn{2}{c|}{\textbf{Task Paradigm}} & \multirow{2}{*}{\textbf{Lane Modeling}} & \textbf{Global Context Supplementation} \\
    \cline{3-4}
     & & Seg & ODet & & \textbf{(Solution for obscure Lanes)} \\
    \midrule
    VPGNet~\cite{VPGNet} & \textit{ICCV'17} & $\uparrow$ & - & Mask & Vanish point prediction \\
    LaneNet~\cite{LaneNet} & \textit{IV'18} & $\uparrow$ & - & Mask & - \\
    SCNN~\cite{SCNN} & \textit{AAAI'18} & $\downarrow$, Max. L. & - & Mask & Spatial CNN layer \\
    EL-GAN~\cite{EL-GAN} & \textit{ECCVW'18} & $\downarrow$, Max. L. & - & Mask & Adversarial training in GAN \\
    Line-CNN~\cite{LineCNN} & \textit{TITS'19} & - & \cmark & Line Anchor & - \\
    SAD~\cite{SAD} & \textit{ICCV'19} & $\downarrow$, Max. L. & - & Mask & Self-attention distillation \\
    FastDraw~\cite{FastDraw} & \textit{ICCV'19} & $\uparrow$ & - & Mask & - \\
    PINet~\cite{PINet} & \textit{TITS'20} & $\uparrow$ & - & Keypoints & - \\
    UFLD~\cite{UFLD} & \textit{ECCV'20} & $\downarrow$, Max. L. & - & Grids & Large FC layer \\
    CurveLane-NAS~\cite{CurveLanes} & \textit{ECCV'20} & - & \cmark & Line Anchor & Feature fusion search module \\
    PolyLaneNet~\cite{PolyLaneNet} & \textit{ICPR'20} & - & \cmark & Polynomial & - \\
    RESA~\cite{RESA} & \textit{AAAI'21} & $\downarrow$, Max. L. & - & Mask & Recurrent feature-shift aggregator \\
    FOLOLane~\cite{FOLOLane} & \textit{CVPR'21} & $\uparrow$ & - & Keypoints & - \\
    LaneATT~\cite{LaneATT} & \textit{CVPR'21} & - & \cmark & Line Anchor & Anchor feature pooling with attention \\
    SGNet~\cite{SGNet} & \textit{IJCAI'21} & - & \cmark & Line Anchor & Perspective attention map \\
    CondLaneNet~\cite{CondLaneNet} & \textit{ICCV'21} & $\downarrow$, Dy. K. & - & Grids & Transformer encoder \\
    LaneAF~\cite{LaneAF} & \textit{RAL'21} & $\uparrow$ & - & Mask & - \\
    LSTR~\cite{LSTR} & \textit{WACV'21} & - & \cmark & Polynomial & Transformer encoder \\
    Laneformer~\cite{Laneformer} & \textit{AAAI'22} & - & \cmark & Line Anchor & Row and column self-attention \\
    GANet~\cite{GANet} & \textit{CVPR'22} & $\uparrow$ & - & Keypoints & Global keypoints association for clustering \\
    Eigenlanes~\cite{Eigenlanes} & \textit{CVPR'22} & - & \cmark & Line Anchor & Pre built candidate set \\
    CLRNet~\cite{CLRNet} & \textit{CVPR'22} & - & \cmark & Line Anchor & ROIGather \\
    B\'{e}zierLaneNet~\cite{BezierLaneNet} & \textit{CVPR'22} & - & \cmark & B\'{e}zier Curve & Feature flip fusion module \\
    UFLD-V2~\cite{UFLDv2} & \textit{TPAMI'22} & $\downarrow$, Max. L. & - & Grids & Large FC layer \\
    RCLane~\cite{RCLane} & \textit{ECCV'22} & $\uparrow$ & - & Keypoints & Global shape message learning \\    
    PGA-Net~\cite{PGANet} & \textit{TITS'23} & - & \cmark & Polynomial & Transformer encoder \\
    PriorLane~\cite{PriorLane} & \textit{ICRA'23} & $\downarrow$, Max. L. & - & Mask & Mixed Transformer as backbone \\
    CondLSTR~\cite{CondLSTR} & \textit{ICCV'23} & $\downarrow$, Dy. K. & - & Keypoints & Dynamic kernels generated by Transformer \\
    ADNet~\cite{ADNet} & \textit{ICCV'23} & - & \cmark & Line Anchor & Large kernel Attention module \\  
    SRLane~\cite{SRLane} & \textit{AAAI'24} & - & \cmark & Line Anchor & Lane segment association module \\
    HGLNet~\cite{HGLNet} & \textit{AAAI'24} & - & \cmark & Line Anchor & Global extraction head via deformable attention \\
    GSENet~\cite{GSENet} & \textit{AAAI'24} & - & \cmark & Line Anchor & Global semantic enhancement module \\
    \bottomrule
    \end{tabular}}
\end{center}
\label{tab:2d_lane_detection_methods}
\vspace{-0.4cm}
\end{table*}

\subsubsection{Keypoints-based modeling}
\label{sec:keypoints}

The mask-based modeling methods often involve predicting numerous irrelevant regions, so some research efforts attempt to directly predict keypoints of lanes. This, like grids-based modeling, can be seen as a sparse version of mask-based modeling, but it directly provides the vectorized expression required by the downstream. 

Some works follow a bottom-up approach. PINet~\cite{PINet} uses a stacked hourglass network to predict keypoint locations and feature embeddings, and clusters different lane instances based on the similarity of feature embeddings. FOLOLane~\cite{FOLOLane} estimates the existence and offset of local lane keypoints, and designs a decoder module with low-level operators that integrates the local information into curve instances. Only keypoints on adjacent boundaries are paired, allowing the network to better focus on detailed features, but the lack of global features causes poor performance under the obscure lanes scene. To add global information, GANet~\cite{GANet} adopts a more efficient post-processing method to cluster points by directly calculating the offset between the keypoint and the start point to globally return to the keypoint. Additionally, a lane-aware feature aggregator based on deformable convolution (DCN)~\cite{DCNv1} is proposed to improve the shape of lanes and better capture the local context on the lane. RCLane~\cite {RCLane} sparsifies the binary segmentation results to obtain the keypoints of all lanes. It decodes the channel in a chained mode using distance and transmission head to predict the keypoints with their continuous relationships in the channel. It also proposes a bilateral prediction method for learning complex topology and global shape information, which can adapt to lanes with complex structures, such as Y-shaped and forked lanes. LanePtrNet~\cite {LanePtrNet} designs a centrality farthest point sampling method to determine the lane center point. Then a grouping head performs clustering based on the center point position and lane point embeddings to obtain the final lane.

Others adopt a top-down manner. For each lane, Chougule et al.~\cite{chougule2018reliable} directly regress the position of keypoints, and Yoo et al.~\cite{E2ELMD} predict a feature map and searches for keypoints of the lane on each row. However, both of them distinguish instances using a multi-classification strategy, as described in Section~\ref{sec:mask}, which is not flexible. CondLSTR~\cite{CondLSTR} improves the instance-obtaining method in CondLaneNet~\cite{CondLaneNet}. It leverages Transformer to generate dynamic and offset kernels for each lane, enhancing global knowledge. Then, these kernels are dynamically convolved with the entire feature map to predict the heatmap and offset maps of lane keypoints.

\subsection{Object Detection based Methods}
\label{sec:object detection-based methods}
\subsubsection{Line Anchor-based modeling}
\label{sec:line anchor}

A lane on the image can be represented by equidistant 2D points. Specifically, the lane is expressed as a sequence of points, i.e., $P = {(x_1, y_1), (x_2, y_2), ..., (x_N, y_N)}$. The y-coordinates of points are equally sampled through the image vertically, i.e., $y_i = \frac{H}{N - 1}\cdot i$, where $H$ is the height of the image. Accordingly, the x-coordinate is associated with the respective $y_i\in Y$. With $P$ and $y_i$, the positions of points that form a lane can be located.

We can initialize a set of two-dimensional points with equal vertical spacing as line anchors. When a line anchor is matched to its corresponding GT, the network only needs to predict the count of valid y-coordinates and the horizontal offset of each valid y-coordinate's x-coordinate relative to the GT. Through this process, the final lane can be reconstructed.

Line-CNN~\cite{LineCNN} uses a large number of predefined straight lines as line anchors. However, it predicts the scores, lengths, and transverse coordinate offsets of all anchors based on the local features of each start point, which implies that the feature map from the backbone must have a sufficiently high number of channels. To solve this problem, LaneATT~\cite{LaneATT} proposes a line anchor feature pooling method that allows the use of a lightweight backbone and presents a high-performance with efficient attention aggregation mechanism to better detect obscure lanes. PointLaneNet~\cite{PointLaneNet} and CurveLane-NAS~\cite{CurveLanes} separate images into non-overlapping grids and regress lanes based on vertical line anchors. In particular, CurveLane-NAS uses network architecture search~\cite{NAS} to find a better network to capture more accurate information, which is beneficial for detecting curved lanes. SGNet~\cite{SGNet} introduces a novel vanish point-oriented anchor generator and adds multiple structural guides to the performance. Jin et al.~\cite{Eigenlanes} introduce data-driven descriptors called eigenlanes, and use lower-order approximations of the lane matrix to obtain line anchors that can better regress curved lanes. Non-maximum suppression (NMS) post-processing is also unavoidable due to the limitations of a large number of predefined anchors and early positive and negative sample matching strategies in object detection.

With the widespread application of Transformers in object detection, research in this field gradually shifts from dense prediction paradigms, such as YOLO~\cite{YOLO} and Faster R-CNN~\cite{FasterRCNN}, to set prediction paradigms like DETR~\cite{DETR}. Similarly, lane detection based on object detection shifts from a fixed dense approach to a dynamic sparse approach. Based on Deformable DETR~\cite{DeformableDETR}, Laneformer~\cite{Laneformer} introduces two novel row and column self-attention operations in the encoder to effectively capture lane context. The binary matching strategy enables an NMS-free approach. Inspired by Sparse R-CNN~\cite{SparseRCNN}, CLRNet~\cite{CLRNet} uses multi-scale feature maps at the pyramid level to iteratively adjust the positions of a small set of preset line anchors~\cite{CascadeRCNN}. It presents ROIGather to enhance lane feature extraction, effectively addressing challenges such as occlusion and lighting variations. Additionally, Line IoU Loss is introduced for global lane regression, enhancing positioning accuracy. CLRNet achieves state-of-the-art results on 2D lane detection datasets. The improvements to Line IoU Loss and label matching are made in~\cite{CLRerNet} and~\cite{CLRmatchNet}, respectively. O2SFormer~\cite{O2SFormer} proposes a one-to-many label allocation strategy and incorporates lane anchor points into position queries~\cite{ConditionalDETR}, providing explicit positional priors that accelerate model convergence. ADNet~\cite {ADNet} removes the limitation of anchor starting points by learning heatmaps for these points and their related directions, enabling the network to adapt to diverse lane types across different datasets. It puts forward a module based on a hybrid CNN\&Transformer architecture~\cite{Conv2Former}~\cite{Segnext} to expand the receptive field, and proposes the Generalized Line IoU Loss to address the limitations of Line IoU Loss~\cite{CLRNet}. Similarly, SRLane~\cite{SRLane} generates sparse line anchors by predicting local directional heatmaps and develops a lane segment association module to adjust non-fitting line anchors. Sparse Laneformer~\cite{SparseLaneformer} designs learnable lane and angle queries to generate sparse line anchors. It employs a two-stage Transformer decoder to refine lane predictions. GSENet~\cite{GSENet} designs a global feature extraction module based on dilated convolution~\cite{Deeplab} and SimAm~\cite{SimAm} to obtain accurate and comprehensive global features, and further enhances semantic representation using ViT~\cite{ViT}. HGLNet~\cite{HGLNet} leverages the large receptive field of dilated convolution to enhance the representation of local features. It designs a global extraction head based on deformable-attention~\cite{DeformableDETR} to extract global feature of lanes adaptively.

\subsubsection{Curve-based modeling}
\label{sec:curve}
Several studies model lanes as curve equations in image space, predicting the parameters of the modeled curves. This idea was first reflected in the works of Gansbeke et al.~\cite{LDE2E}. They propose a differentiable least squares fitting module, which fits the cubic polynomial curves (e.g. \(x=ay^3+by^2+cy+d\)) to the points predicted by deep neural networks. Then, PolyLaneNet~\cite{PolyLaneNet} directly learns to predict polynomial coefficients with simple fully connected layers. LSTR~\cite{LSTR} uses Transformer to predict polynomials in an end-to-end manner based on DETR~\cite{DETR}. However, the performance of these methods remains suboptimal due to the difficulty of the curve's parameter learning and challenges in transformer training. PGA-Net~\cite{PGANet} introduces an improved supervised strategy to accelerate transformer convergence and proposes a Mean Curvature Loss to constrain the curvature of predicted lanes, enhancing the predictive accuracy for curved lanes.

Feng et al.~\cite{BezierLaneNet} argue that the polynomials are abstract and the coefficients are challenging to optimize, recommending third-order B\'{e}zier curves for lane modeling. Their network predicts four control points to determine lane positions, proving more robust than direct regression of polynomial coefficients. They also consider the pseudo-symmetry of lanes in images and propose a feature-flipping fusion module based on DCN~\cite{DCNv2} to enhance feature representation in vehicle front-view images. Subsequently, Chen et al.~\cite{BSNet} model lanes as more flexible B-spline curves and propose a novel curve-distance calculation method to improve control points prediction supervision.

\subsection{2D Lanes to 3D Lanes}
\label{sec:IPM}
Once we obtain the 2D lane coordinates from FV, we need to use IPM to project it into BEV to obtain the 3D lanes for downstream use. We briefly review the general IPM process here. Firstly, the relationship between each pixel coordinates $(u, v)$ and camera coordinates $(x_c, y_c, z_c)$ can be described as:

\begin{equation}\label{pixel-camera}
z_c \begin{bmatrix} u \\ v \\ 1 \\ \end{bmatrix}
= \bm{K} \begin{bmatrix} x_c \\ y_c \\ z_c \\ \end{bmatrix},
\end{equation}

where the matrix $\bm{K}$ represents the camera intrinsics. Then each camera coordinates and ego vehicle coordinates $(x_e, y_e, z_e)$ can be linked as:

\begin{equation}\label{camera-ego}
\begin{bmatrix} x_c \\ y_c \\ z_c \end{bmatrix}
= \bm{R} \begin{bmatrix} x_e \\ y_e \\ z_e \\ \end{bmatrix} + \bm{T},
\end{equation}

where $\bm{R}$, $\bm{T}$ refers to a rotation and a translation matrix, respectively. Their combination denotes the camera extrinsics.

With Eqn.~\ref{pixel-camera} and Eqn.~\ref{camera-ego}, we can establish a transformation from each pixel coordinates to ego vehicle coordinates:
\begin{equation}\label{pixel-ego}
\begin{aligned}
\begin{bmatrix} x_e \\ y_e \\ z_e \\ \end{bmatrix}
& = z_c \bm{R}^{-1} \bm{K}^{-1} \begin{bmatrix} u \\ v \\ 1 \\ \end{bmatrix} - \bm{R}^{-1} \bm{T}. \\
\end{aligned}
\end{equation}

Due to the characteristics of perspective projection, objects in 3D space may lose depth information when imaged by the camera onto the image plane. It means that objects at different distances may be projected onto the same position. So when we only have $(u, v)$ and camera intrinsics and extrinsics, we cannot obtain $(x_e, y_e, z_e)$. We have to assume that the ground is flat, i.e. $z_e$ is a constant. Let \(\bm{M}_1 = \bm{R}^{-1}K^{-1}\begin{bmatrix} u & v & 1 \end{bmatrix}^T\) and \(\bm{M}_2 = \bm{R}^{-1}\bm{T}\), then according to Eqn.~\ref{pixel-ego}, $z_c$ can be calculated as:

\begin{equation}\label{z_c}
z_c = \frac{z_e + \bm{M}_2(2, 0)}{\bm{M}_1(2, 0)}. \\
\end{equation}

Finally, substituting Eqn.~\ref{z_c} into Eqn.~\ref{pixel-ego} can obtain the ego vehicle coordinates. It is cumbersome to transform every pixel coordinate according to this manner. We can select four points on FV as Region Of Interest (ROI) and calculate the corresponding positions in the ego vehicle coordinate system using the above method. Then we can establish a system of ternary linear equations:
\begin{equation}\label{H_ipm}
\begin{aligned}
\begin{bmatrix} x_e \\ y_e \\ z_e \\ \end{bmatrix} = \bm{H} \begin{bmatrix} u \\ v \\ 1 \\ \end{bmatrix} = \begin{bmatrix} a_{11} & a_{12} & a_{13} \\ a_{21} & a_{22} & a_{23} \\ a_{31} & a_{32} & a_{33} \\ \end{bmatrix} \begin{bmatrix} u \\ v \\ 1 \\ \end{bmatrix}. \\
\end{aligned}
\end{equation}

By using the known four point pairs, we can solve the inverse perspective transformation matrix $\bm{H}$. Then we can use $\bm{H}$ to obtain the position of ROI corresponding to the ego vehicle coordinate system in the image. 

\begin{figure}
\centering
\includegraphics[width=0.48 \textwidth]{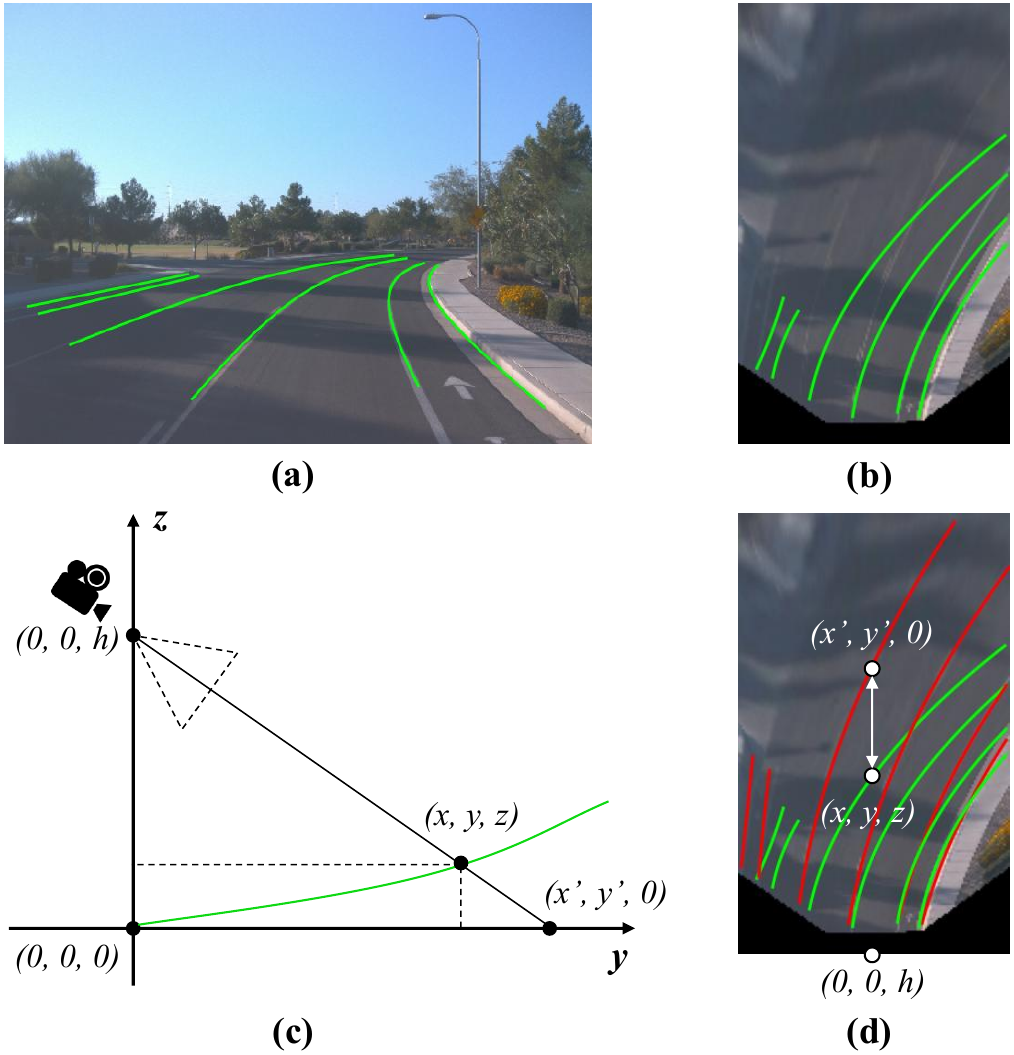}
\caption{
Illustration of the errors introduced by IPM on the uphill path. (a) The GT of 2D lanes in the FV image. (b) The GT of 3D lanes on the virtual BEV plane. (c) The co-linear relationship between a 3D lane point $(x, y, z)$, its projection $(x^{'}, y^{'}, 0)$ on the virtual BEV plane and camera center $(0, 0, h)$. (d) Comparison of lanes obtained through IPM projection with GT on the virtual BEV plane. Due to the assumption of flat ground, when the vehicle goes uphill, the 3D lanes obtained by IPM are divergent rather than parallel. Similarly, they converge when going downhill.
}
\label{fig:ipm_error}
\vspace{-0.4cm}
\end{figure}

Although there are better IPM processes available~\cite{1998IPM, 2014IPM, 2016IPM}, the assumption of flat ground is inevitable due to the perspective effect. As shown in Figure~\ref{fig:ipm_error}, the lanes would diverge/converge during uphill/downhill, potentially leading to improper action decisions in the planning and control module if the height is ignored. This is why there has been a focus on directly predicting 3D lanes from FV~\cite{PersFormer, BEVLaneDet, Anchor3DLane, LATR}.

\begin{figure*}[t]
\centering
\includegraphics[width=1 \textwidth]{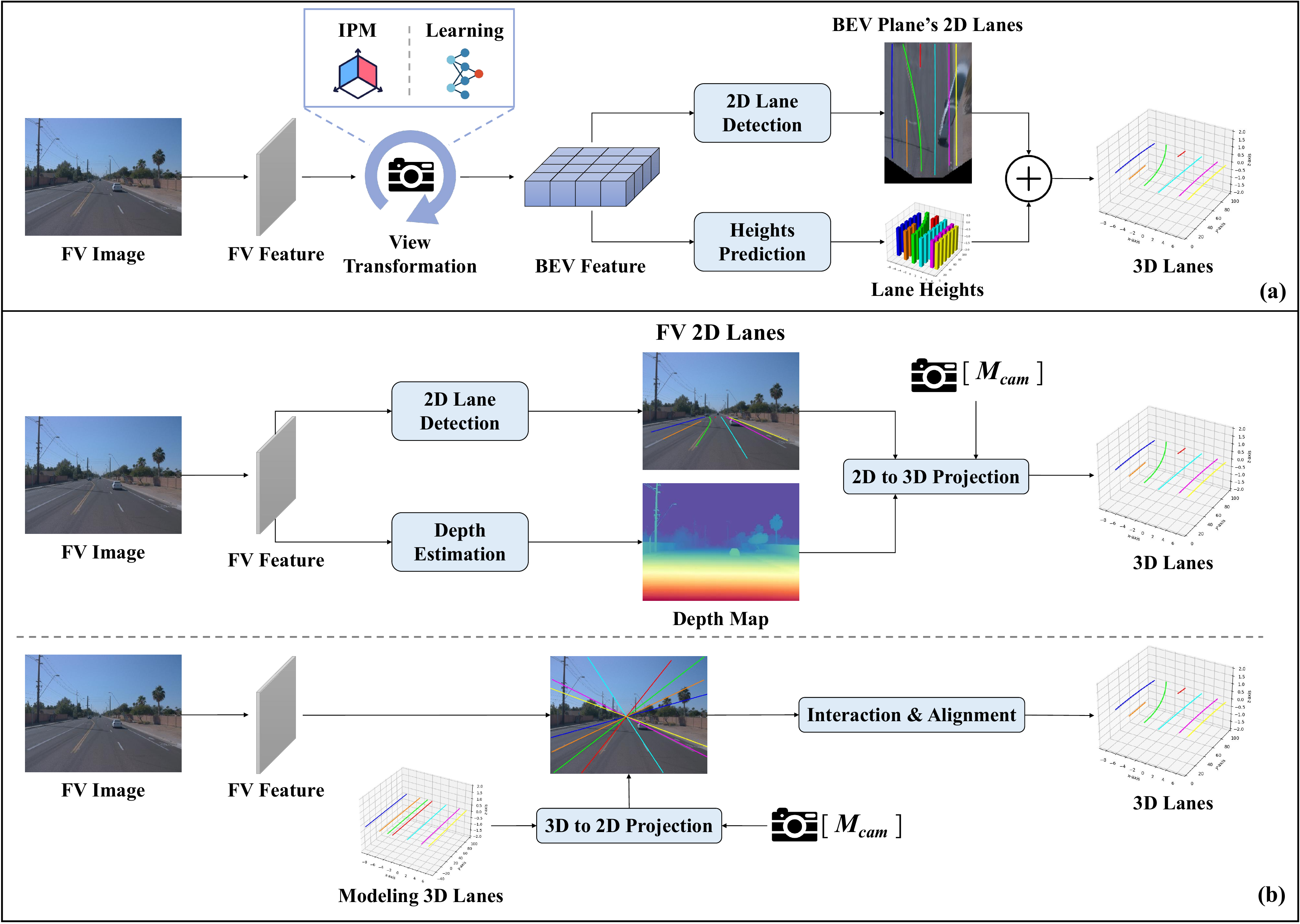}
\caption{
The general pipeline of 3D lane detection. (a) BEV-based methods. The core is the view transformation from FV features to BEV features, including IPM and learning approach. (b) BEV-free methods. There are two branches: one is to project 2D lanes into 3D space based on depth estimation results, and the other is to directly model 3D lanes and project them back into FV for interaction and alignment.
}
\label{fig:methods_3d}
\vspace{-0.4cm}
\end{figure*}

\section{Methods of 3D Lane Detection}
\label{sec:methods_3d}
This section reviews recent 3D lane detection methods. We first explain the ground for classifying existing methods in Section~\ref{sec:taxonomy_3d}, and then discuss the existing methods accordingly in Section~\ref{sec:bev_based_methods} and Section~\ref{sec:bev_free_methods}. 

\subsection{Classification Framework}
\label{sec:taxonomy_3d}
As an upgrade to 2D lane detection, 3D lane detection primarily focuses on how to utilize neural networks to reconstruct the missing 3D information from 2D FV images. 

As shown in Figure~\ref{fig:methods_3d}, existing 3D lane detection methods can be divided into two categories: (a) \textbf{BEV-based methods}, which utilize camera parameters and convert the extracted FV features into BEV features with height information in some way. This process of constructing an intermediate proxy is usually referred to as view transformation~\cite{BEVsurvey}. In this way, the 3D lane detection task can be simplified to 2D lane detection in BEV, and then combining it with the corresponding height values estimated by a height estimation head yields the final three-dimensional lanes. Therefore, the performance of this type of method depends not only on the 2D lane detection results in BEV but also on the adopted view transformation method. (b) \textbf{BEV-free methods}, which do not hinge on BEV features. It can be further divided into two types. One is to detect 2D lanes in the FV image while predicting their depth, and then project them onto the 3D space. The other is to directly model lanes in the 3D space. With the initialized 3D information, it is possible to project it onto FV based on camera parameters. This approach enables direct interactions between the 3D lane and FV features, ultimately refining and updating the 3D lane.

Under the classification framework, for each specific method, instance discrimination and lane modeling are still discussed. We list a comparison of representative works in Table~\ref{tab:3d_lane_detection_methods}. 

\begin{table*}[ht]
\scriptsize
\renewcommand\arraystretch{1.2}
\centering
    \caption{The summary of representative 3D lane detection methods. For BEV-based methods, the specific implementation of view transformation is additionally indicated instead of using "\cmark" to mark. For \textbf{Task Paradigm}, "Seg" represents segmentation-based methods, "ODet" represents object detection-based methods. In segmentation-based methods, the approach for instance discrimination is indicated instead of marking with "\cmark". "↑" indicates the bottom-up manner for instance discrimination.}
\begin{center}
    \resizebox{1.0\textwidth}{!}{
    \begin{tabular}{l|c|c|c|c|c|c} 
    \toprule
    \multirow{2}{*}{\textbf{Methods}} & \multirow{2}{*}{\textbf{Venue}} &
    \multicolumn{2}{c|}{\textbf{Perspective Effect Elimination}} & \multicolumn{2}{c|}{\textbf{Task Paradigm}} & \multirow{2}{*}{\textbf{Lane Modeling}}
    \\
    \cline{3-6}
     & & BEV-based & BEV-free & Seg & ODet & \\
    \midrule
    3D-LaneNet~\cite{3DLaneNet} & \textit{ICCV'19} & IPM & - & - & \cmark & Line Anchor \\
    Gen-LaneNet~\cite{GenLaneNet} & \textit{ECCV'20} & IPM & - & - & \cmark & Line Anchor \\ 
    CLGo~\cite{GenLaneNet} & \textit{AAAI'22} & IPM & - & - & \cmark & Polynomial \\ 
    SALAD~\cite{ONCE3DLanes} & \textit{CVPR'22} & - & \cmark & $\uparrow$ & - & Mask \\
    Reconstruct~\cite{li2022reconstruct} & \textit{CVPRW'22} & IPM & - & - & \cmark & Line Anchor \\
    PersFormer~\cite{PersFormer} & \textit{ECCV'22} & PersFormer & - & - & \cmark & Line Anchor \\
    3D-SplineNet~\cite{3DSplineNet} & \textit{WACV'23} & IPM & & - & \cmark  & B-Spline Curve \\
    CurveFormer~\cite{CurveFormer} & \textit{ICRA'23} & - & \cmark & - & \cmark & 3D Line Anchor \\    
    Anchor3DLane~\cite{Anchor3DLane} & \textit{CVPR'23} & - & \cmark& - & \cmark & 3D Line Anchor \\
    BEV-LaneDet~\cite{BEVLaneDet} & \textit{CVPR'23} & VPN~\cite{VPN} & - & $\uparrow$ & - & Keypoints \\ 
    SPG3DLane~\cite{SPG3DLane} & \textit{ICCV'23} & PersFormer & - & - & \cmark & Line Anchor \\
    LATR~\cite{LATR} & \textit{ICCV'23} & - & \cmark & - & \cmark & 3D Line Anchor \\
    PVALane~\cite{PVALane} & \textit{AAAI'24} & - & \cmark & - & \cmark& 3D Line Anchor \\
    B\'{e}zierFormer~\cite{BezierFormer} & \textit{ICME'24} & - & \cmark & - & \cmark & 3D B\'{e}zier Curve \\
    LaneCPP~\cite{LaneCPP} & \textit{CVPR'24} & LSS~\cite{LSS} & - & - & \cmark & B-Spline Curve \\
    \bottomrule
    \end{tabular}}
\end{center}
\label{tab:3d_lane_detection_methods}
\vspace{-0.4cm}
\end{table*}

\subsection{BEV-based methods}
\label{sec:bev_based_methods}
The pipeline follows the established process of dense BEV perception methods~\cite{LSS, BEVDet, BEVFormer}. Since BEV features inherently conceal height information, the subsequent lane decoding process only needs to consider the 2D BEV plane, which can naturally be integrated with 2D lane detection methods. The view transformation between FV and BEV features can be formulated as:
\begin{equation}\label{BEVencoder}
F_{BEV}(x, y, z) = M_{trans}(F_{FV}(u, v), [\bm{R \quad T}], \bm{K}),
\end{equation}
where $F_{FV}$ denotes the FV feature. $F_{BEV}$ denotes the BEV feature which contains height information. $x$, $y$, $z$ denote coordinates in 3D space. $M_{trans}$ denotes view transformation module. $u$, $v$ denote corresponding pixel coordinates in terms of $x$, $y$, $z$. $[\bm{R \quad T}]$ and $\bm{K}$ are camera extrinsics and intrinsics.

\subsubsection{IPM for View Transformation}
\label{sec:ipm_view_trans}
3D-LaneNet~\cite{3DLaneNet} is the first method which uses deep learning to predict 3D lanes directly from monocular images. The network first predicts the camera pitch angle and height to generate a differentiable IPM, combining the original FV feature map to create the BEV feature map. 3D-LaneNet demonstrates promising results in detecting 3D lanes from monocular images. Then Gen-LaneNet~\cite{GenLaneNet} directly uses 2D lane segmentation results as input for IPM, allowing for the utilization of extensive 2D lane data and enhancing the model's generalization. In contrast to 3D-LaneNet, Gen-LaneNet offers more reliable supervision by using the camera pitch angle and height as GT. Li et al.~\cite{li2022reconstruct} propose a new loss function based on Gen LaneNet to better extract the height information of 3D lanes from 2D lane representations. On the BEV plane, the above work uses vertical line anchors to model lanes. However, matching line anchors to ground truth is performed by measuring the distance at a predefined $y = Y_{ref}$, which may result in missed detections for short lanes. Therefore, 3D-LaneNet+~\cite{3DLaneNet+} avoids this issue using a bottom-up segmentation approach. Liu et al.~\cite{CLGo} believe that the model can be independent of the ground truth camera pose provided by the benchmark. They design a two-stage network based on Transformer, which first predicts camera pose, i.e., the required parameters for IPM, then extracts BEV features, and finally regresses polynomial coefficients. 3D-SplineNet~\cite{3DSplineNet} treats lanes as B-spline curves on the BEV plane.

These methods offer valuable guidance for the initial exploration of 3D lane detection. However, their acquisition of BEV features relies on IPM. As discussed in Sec~\ref{sec:IPM}, this rigid mapping lacks robustness. Issues such as improper feature transformation and suboptimal performance tend to arise during bending or squeezing turns.

\subsubsection{Learnable View Transformation}
\label{sec:learn_view_trans}
To reduce the inherent errors caused by this rigid transformation, some researchers adopt a more flexible approach. They use neural networks to learn the transformation process from FV to BEV features. PersFormer~\cite{PersFormer} leverages deformable attention~\cite{DeformableDETR} to learn the spatial transformation from FV to BEV. It references the coordinate transformation matrix of IPM to generate BEV feature representations, focusing on relevant regions in the FV features. In the lane decoding stage, PersFormer adopts a unified 2D/3D line anchor design, achieving unified 2D and 3D lane detection. BEV-LaneDet~\cite{BEVLaneDet} integrates the MLP based view transformation method VPN~\cite{VPN} into FPN~\cite{FPN} to obtain BEV features. It constructs a virtual camera module to project all images onto a standard virtual camera view, ensuring consistent image distribution. For lane modeling, BEV-LaneDet models lanes as keypoints. It adopts an embedding-clustering-based instance segmentation method~\cite{LaneNet} and refers to YOLO~\cite{YOLO} to divide the BEV plane into grids, predicting the offset of each grid's center point relative to GT. Due to its concise architecture, it is well-suited for deployment. Yao et al.~\cite{SPG3DLane} add the coarse-to-fine mechanism~\cite{CLRNet} based on PersFormer~\cite{PersFormer}. They fuse the local and global information referring to the coordinates of sparse points and jointly refine the global and local structures of lanes. Chen et al.~\cite{EfficientTR} decompose the cross-attention between FV and BEV features into separate cross-attentions: one between FV and lane features, and another between BEV and lane features. Dynamic kernels are then used to convolve FV and BEV feature maps, generating 2D and 3D lane keypoint offset maps. GroupLane~\cite{GroupLane} uses the depth-estimation-based view transformation method LSS~\cite{LSS} to obtain BEV features. It models BEV lanes as grids~\cite{UFLD} and establishes vertical and horizontal group heads to identify horizontal and vertical lanes, respectively. LaneCPP~\cite{LaneCPP} also completes BEV transformation based on LSS. It models lanes as B-spline curves and uses prior knowledge of road geometry to enhance view transformation and lane prediction.

\subsection{BEV-free Methods}
\label{sec:bev_free_methods}

\subsubsection{Combining Depth Estimation}
\label{sec:depth_estimation}
Similar to depth-assisted methods~\cite{SMOKE, FCOS3D, MonoDLE} in monocular 3D object detection, SALAD~\cite{ONCE3DLanes} decouples 3D lane detection into 2D lane segmentation and dense depth estimation tasks. With the help of estimated depth, the 2D lane coordinates can be projected into 3D space. Due to the availability of depth information, this method is independent of camera extrinsics.

\begin{table*}[ht]
\centering
\caption{Benchmark results of the representative 2D lane detection methods on CULane~\cite{SCNN}. For “Cross” scene, only false positives are shown. For convenient comparison, a boundary line is used to separate segmentation-based methods and object detection-based methods.}
\begin{center}
\resizebox{1.0\textwidth}{!}{\begin{tabular}{lccccccccccc}
\toprule
\textbf{Method} & \textbf{Backbone} & \textbf{F1($\%$)}  & \textbf{Normal($\%$)} & \textbf{Crowded($\%$)} & \textbf{Dazzle($\%$)} & \textbf{Shadow($\%$)} & \textbf{No line($\%$)}  & \textbf{Arrow($\%$)} & \textbf{Curve($\%$)} & \textbf{Cross} & \textbf{Night($\%$)} \\
\midrule
SCNN~\cite{SCNN} & VGG16 & 71.60 & 90.60 & 69.70 & 58.50 & 66.90 & 43.40 & 84.10 & 64.40 & 1990 & 66.10 \\
SAD~\cite{SAD} & ResNet101 & 71.80 & 90.70 & 70.00 & 59.90 & 67.00  & 43.50 & 84.40 & 65.70 & 2183 & 65.90 \\
RESA~\cite{RESA} & ResNet34 & 74.50 & 91.90 & 72.40 & 66.50 & 72.00  & 46.30 & 88.10 & 68.60 & 1896 & 69.80 \\
LaneAF~\cite{LaneAF} & ERFNet & 75.63 & 91.10 & 73.32 & 69.71 & 75.81 & 50.62 & 86.86 & 65.02 & 1844 & 70.90 \\
UFLD~\cite{UFLD} & ResNet34 & 72.30 & 90.70 & 70.20 & 59.50 & 69.30 & 44.40 & 85.70 & 69.50 & 2037 & 66.70 \\
CondLaneNet~\cite{CondLaneNet} & ResNet18 & 78.14  & 92.87 & 75.79 & 70.72 & 80.01 & 52.39 & 89.37  & 72.40 & 1364 & 73.23 \\
UFLDv2~\cite{UFLDv2} & ResNet18 & 74.70 & 91.70 & 73.00 & 64.60 & 74.70 & 47.20 & 87.60 & 68.70 & 1998 & 70.20 \\
PINet-4H~\cite{PINet} & - & 74.40 & 90.30 & 72.30 & 66.30 & 68.40 & 49.80 & 83.70 & 65.20 & 1427 & 67.70 \\
FOLOLane~\cite{FOLOLane} & ERFNet & 78.80 & 92.70 & 77.80 & 75.20 & 79.30 & 52.10 & 89.00 & 69.40 & 1569 & 74.50 \\
GANet~\cite{GANet} & ResNet18 & 78.79 & 93.24 & 77.16 & 71.24 & 77.88 & 53.59 & 89.62 & 75.92 & 1240 & 72.75 \\
RCLane~\cite{RCLane} & SegFormerB0 & 79.52 & 93.41  & 77.93 & 73.32 & 80.31 & 53.84 & 89.04 & 75.66 & 1298 & 74.33 \\
CondLSTR~\cite{CondLSTR} & ResNet18 & 80.36 & 94.11 & 79.17 & 73.55 & 80.39 & 54.41 & 90.37 & 75.89 & 1214 & 75.39 \\
\midrule
CurveLane-L~\cite{CurveLanes} & - & 74.80 & 90.70 & 72.30 & 67.70 & 70.10 & 49.40 & 85.80 & 68.40 & 1746 & 68.90 \\
LaneATT~\cite{LaneATT} & ResNet18 & 75.09 & 91.11 & 72.96 & 65.72 & 70.91 & 48.35 & 87.82 & 63.37 & 1170 & 68.95 \\
SGNet~\cite{SGNet} & ResNet18 & 76.12 & 91.42 & 74.05 & 66.89 & 72.17 & 50.16 & 87.13 & 67.02 & 1164 & 70.67 \\
Eigenlanes~\cite{Eigenlanes} & ResNet18 & 76.50 & 91.50 & 74.80 & 69.70 & 72.30 & 51.10 & 87.70 & 62.00 & 1507 & 71.40 \\
CLRNet~\cite{CLRNet} & ResNet18 & 79.58 & 93.30 & 78.33 & 73.71 & 79.66 & 53.14 & 90.25 & 71.56 & 1321 & 75.11 \\
ADNet~\cite{ADNet} & ResNet18 & 77.56 & 91.92 & 75.81 & 69.39 & 76.21 & 51.75 & 87.71 & 68.84 & 1133 & 72.33 \\
B\'{e}zierLaneNet~\cite{BezierLaneNet} & ResNet18 & 73.67 & 90.22 & 71.55  & 62.49 & 70.91 & 45.30 & 84.09 & 58.98 & 996  & 68.70 \\
PGA-Net~\cite{PGANet} & ResNet18 & 69.86 & 87.84 & 70.00 & 62.11 & 67.61 & 46.71 & 80.94 & 58.01 & 1700 & 59.02 \\
\bottomrule
\end{tabular}}
\end{center}
\label{tab:culane_result}
\vspace{-0.4cm}
\end{table*}

\subsubsection{Derictly Modeling 3D Lanes}
\label{sec:model_3d_lane}
The 3D object detection methods based on sparse BEV representations, such as DETR3D~\cite{DETR3D} and PETR~\cite{PETR}, guide this approach. CurveFormer~\cite{CurveFormer} constructs curve queries by modeling 3D lanes as 3D line anchors to provide explicit positional priors~\cite{DABDETR}. It designs a curve cross-attention mechanism to predict polynomial parameters for 3D lanes. Inspired by the line anchor feature pooling mechanism in LaneATT~\cite{LaneATT}, Anchor3DLane~\cite{Anchor3DLane} utilizes camera intrinsics and extrinsics to accurately project 3D line anchor points onto FV features. This facilitates anchor feature sampling, allowing the network to predict 3D coordinates and lane classification results based on the sampled anchor features. LATR~\cite{LATR} decomposes 3D line anchors into dynamically generated point-level and lane-level queries. It uses dynamic 3D ground position embeddings to interact with FV features, updating the lane query to bridge 3D space and 2D images. Dong et al.~\cite{BezierFormer} model lanes as 3D B\'{e}zier curves and predict the curve control points via Transformer. Han et al.~\cite{DecoupleLane} express the 3D lane as a polynomial in 3D space. They design two Transformer structures to learn the 2D polynomial with the height of the X-O-Z plane and project the resulting 3D lane onto FV for supervised alignment. PVALane~\cite{PVALane} generates sparser 3D line anchors than Anchor3DLane~\cite{Anchor3DLane} by predicting 2D lanes in FV. It also introduces a module to align sampled FV and BEV features for more accurate 3D lane detection.

\section{Benchmark Results}
\label{sec:benchmark}
This section reports the performance of representative lane detection methods on commonly used public datasets. For each reviewed area, the most widely used datasets are selected for benchmarking in Section~\ref{sec:main_results}. Because of the high efficiency requirement for lane detection, speed tests are also conducted on representative open-source lane detection methods in a unified environment which are shown in Section~\ref{sec:efficiency}. Note that we only list published works for reference. Following the performance and efficiency comparisons, Section~\ref{sec:discussion} revisits the existing methods according to the four core designs of lane detection.

\subsection{Main Results on Lane Detection Datasets}
\label{sec:main_results}
CULane~\cite{SCNN} and OpenLane~\cite{PersFormer} are currently the most widely used 2D and 3D lane detection datasets. We report the performance of representative methods on these two benchmarks in Table~\ref{tab:culane_result} and Table~\ref{tab:openlane_result} separately. All results are derived from the data in the original paper. More benchmark results are reported in Section~\ref{sec:more_benchmark_results} of Appendix.

\subsection{Efficiency Comparison}
\label{sec:efficiency}
Since the different methods are implemented on different platforms for the experiment, it is unfair to directly compare the speeds reported in their original papers. Therefore, we retest representative methods in a unified environment. Table~\ref{tab:efficiency} shows the work efficiency of these methods. The representative open-source methods are reevaluated according to their settings on the CULane or OpenLane dataset. To ensure fairness, only the inference speed of the model is tested to report the frames per second (FPS). The backbone, input size, model's output, and possible post-processing (whether the model's output reflects a vectorized representation of each unique lane instance) of each method are also described. All tests are conducted on a single Nvidia GeForce RTX 3090 GPU.

\begin{table*}[t]
\centering
\caption{Benchmark results of the representative 3D lane detection methods on OpenLane~\cite{PersFormer}.}
\renewcommand{\arraystretch}{1.2}
\resizebox{1.0\textwidth}{!}{\begin{tabular}{lcccccccc ccccc}
\toprule
\multirow{2}{*}{\textbf{Method}} & \multirow{2}{*}{\textbf{Backbone}} & \multirow{2}{*}{\textbf{F1($\%$)}} & \textbf{Up \&} & \multirow{2}{*}{\textbf{Curve($\%$)}} & \textbf{Extreme} & \multirow{2}{*}{\textbf{Night($\%$)}} & \multirow{2}{*}{\textbf{Intersection($\%$)}} & \textbf{Merge \&} & \textbf{Cate} & \multicolumn{2}{c}{\textbf{X error(m)}} & \multicolumn{2}{c}{\textbf{Z error(m)}} \\ 
\cmidrule(lr){11-12} \cmidrule(lr){13-14}
 & & & \textbf{Down($\%$)} & & \textbf{Weather($\%$)}& & & \textbf{Split($\%$)} & \textbf{Acc($\%$)} & \textbf{near} & \textbf{far} & \textbf{near} & \textbf{far} \\
\midrule
3D-LaneNet~\cite{3DLaneNet} & VGG16 & 44.1 & 40.8 & 46.5 & 47.5 & 41.5 & 32.1 & 41.7 & - & 0.479 & 0.572 & 0.367 & 0.443 \\ 
GenLaneNet~\cite{GenLaneNet} & ERFNet & 32.3 & 25.4 & 33.5 & 28.1 & 18.7 & 21.4 & 31.0 & - & 0.593 & 0.494 & 0.140 & 0.195 \\ 
PersFormer~\cite{PersFormer} & EfficientNetB7 & 50.5 & 42.4 & 55.6 & 48.6 & 46.6 & 40.0 & 50.7 & 89.5 & 0.319 & 0.325 & 0.112 & 0.141 \\ 
CurveFormer~\cite{CurveFormer} & EfficientNetB7 & 50.5 & 45.2 & 56.6 & 49.7 & 49.1 & 42.9 & 45.4 & - & 0.340 & 0.772 & 0.207 & 0.651 \\ 
Anchor3DLane~\cite{Anchor3DLane} & EfficientNetB3 & 56.0 & 50.3 & 59.1 & 53.6 & 52.8 & 47.4 & 53.3 & 89.9 & 0.293 & 0.317 & 0.103 & 0.130 \\ 
BEV-LaneDet~\cite{BEVLaneDet} & ResNet34 & 58.4 & 48.7 & 63.1 & 53.4 & 53.4 & 50.3 & 53.7 & - & 0.309 & 0.659 & 0.244 & 0.631 \\ 
SPG3DLane~\cite{SPG3DLane} & EfficientNetB7 & 53.7 & 46.2 & 59.2 & 54.8 & 49.8 & 41.9 & 52.1 & - & 0.468 & 0.514 & 0.371 & 0.418 \\
LATR~\cite{LATR} & ResNet50 & 61.9 & 55.2 & 68.2 & 57.1 & 55.4 & 52.3 & 61.5 & 92.0 & 0.219 & 0.259 & 0.075 & 0.104 \\
PVALane~\cite{PVALane} & ResNet50 & 62.7 & 54.1 & 67.3 & 62.0 & 57.2 & 53.4 & 60.0 & 93.4 & 0.232 & 0.259 & 0.092 & 0.118 \\
LaneCPP~\cite{LaneCPP} & EfficientNetB7 & 60.3 & 53.6 & 64.4 & 56.7 & 54.9 & 52.0 & 58.7 & - & 0.264 & 0.310 & 0.077 & 0.117 \\
\bottomrule
\end{tabular}}
\label{tab:openlane_result}
\end{table*}

\begin{table*}[ht]
\centering
\small
\caption{Efficiency comparison of representative methods. The setting of SCNN~\cite{SCNN} and LSTR~\cite{LSTR} is based on the re-implementation of Feng et al.~\cite{BezierLaneNet}. 3D-LaneNet~\cite{3DLaneNet} is not open-source, so we tested it according to Guo et al.'s reproduction~\cite{GenLaneNet}. For \textbf{Method}, ''$\dagger$'' denotes the iterative regression of Anchor3DLane~\cite{Anchor3DLane} and ''LATR-Lite'' refers to the lite version of LATR~\cite{LATR}.}
\renewcommand{\arraystretch}{1.3}
\setlength{\tabcolsep}{3.0pt}
\scalebox{0.67}{
\begin{tabular} {l|p{0.15\textwidth}<{\centering}|p{0.1\textwidth}<{\centering}|p{0.05\textwidth}<{\centering}|p{0.67\textwidth}<{\centering}|p{0.28\textwidth}<{\centering}}
\toprule
\belowrulesepcolor{gray!30!}
Method & Backbone & Input Size & FPS & Output & Post-processing \\  
\midrule
\belowrulesepcolor{gray!15!}
\rowcolor{gray!15!}\multicolumn{6}{c}{\textbf{2D Lane Detection Methods (Section~\ref{sec:methods_2d})}} \\ 
\aboverulesepcolor{gray!15!} 
\midrule
SCNN~\cite{SCNN} & ResNet18 & 288$\times$800 & 14 & Multi-classes semantic segmentation mask. & Vectorization \\ 
\midrule
SAD~\cite{SAD} & ERFNet & 208$\times$976 & 92 & Multi-classes semantic segmentation mask. & Vectorization \\ 
\midrule
RESA~\cite{RESA} & ResNet18 & 288$\times$800 & 68 & Multi-classes semantic segmentation mask. & Vectorization  \\ 
\midrule
LaneAF~\cite{LaneAF} & ERFNet & 288$\times$832 & 63 & Binary segmentation mask and affinity vector fields. & Clustering $\&$ Vectorization \\ 
\midrule
UFLD~\cite{UFLD} & ResNet18 & 288$\times$800 & 358 & Multi classification probability of grid for each row. & Vectorization \\ 
\midrule
LSTR~\cite{LSTR} & ResNet18 & 288$\times$800 & 133 & Each cubic polynomial's classification probability and coefficient values. & None \\
\midrule
LaneATT~\cite{LaneATT} & ResNet18 & 360$\times$640 & 194 & Lane's classification probability and equidistant point coordinates. & NMS \\
\midrule
CondLaneNet~\cite{CondLaneNet} & ResNet18 & 320$\times$800 & 219 & Multi classification probability of grid for each row. & Vectorization \\ 
\midrule
B\'{e}zierLaneNet~\cite{BezierLaneNet} & ResNet18 & 288$\times$800 & 244 & Each cubic B\'{e}zier curve‘s classification probability and control point coordinates. & None \\
\midrule
GANet~\cite{GANet} & ResNet18 & 320$\times$800 & 106 & Keypoint heatmaps and offset maps of all lanes foreground. & Clustering \& Coordinates refinement \\
\midrule
CLRNet~\cite{CLRNet} & ResNet18 & 320$\times$800 & 104 & Lane's classification probability and equidistant point coordinates. & NMS \\
\midrule
CondLSTR~\cite{CondLSTR} & ResNet18 & 320$\times$800 & 47 & Keypoint heatmaps and offset maps of each lane instance. & Coordinates refinement \\
\midrule
ADNet~\cite{ADNet} & ResNet18 & 320$\times$800 & 109 & Lane's classification probability and equidistant point coordinates. & NMS \\
\midrule
\belowrulesepcolor{gray!15!}
\rowcolor{gray!15!}\multicolumn{6}{c}{\textbf{3D Lane Detection Methods (Section~\ref{sec:methods_3d})}} \\ \aboverulesepcolor{gray!15!} \midrule
3D-LaneNet~\cite{3DLaneNet} & VGG16 & 360$\times$480 & 118 & BEV lane's classification probability and equidistant point coordinates, 3D lane heights. & NMS \\ 
\midrule
Gen-LaneNet~\cite{GenLaneNet} & ERFNet & 360$\times$480 & 24 & BEV lane's classification probability and equidistant point coordinates, 3D lane heights. & NMS \\
\midrule
PersFormer~\cite{PersFormer} & EfficientNetB7 & 360$\times$480 & 19 & BEV lane's classification probability and equidistant point coordinates, 3D lane heights. & NMS \\  
\midrule
PersFormer~\cite{PersFormer} & EfficientNetB7 & 720$\times$960 & 12 & BEV lane's classification probability and equidistant point coordinates, 3D lane heights. & NMS \\  
\midrule
Anchor3DLane~\cite{Anchor3DLane} & ResNet18 & 360$\times$480 & 75 & 3D lane's classification probability and equidistant point coordinates. & NMS \\ 
\midrule
Anchor3DLane$\dagger$~\cite{Anchor3DLane} & ResNet50 & 720$\times$960 & 18 & 3D lane's classification probability and equidistant point coordinates. & NMS \\ 
\midrule
BEV-LaneDet~\cite{BEVLaneDet} & ResNet34 & 576$\times$1024 & 83 & BEV lane's keypoint coordinates and instance embedding, 3D lane heights. & Clustering \\ 
\midrule
SPG3DLane~\cite{SPG3DLane} & EfficientNetB7 & 720$\times$960 & 13 & BEV lane's classification probability and equidistant point coordinates, 3D lane heights. & NMS \\ 
\midrule
LATR~\cite{LATR} & ResNet50 & 720$\times$960 & 14 & 3D lane classification probability and equidistant point coordinates. & None \\ 
\midrule
LATR-Lite~\cite{LATR} & ResNet50 & 720$\times$960 & 22 & 3D lane classification probability and equidistant point coordinates. & None \\ 
\midrule
\end{tabular}
}
\label{tab:efficiency}
\vspacefigtext
\end{table*}

\subsection{Discussion}
\label{sec:discussion}
In the two previous chapters, the overview of existing methods is presented from four aspects: task paradigm, lane modeling, global context supplementation, and perspective effect elimination. Combining performance and efficiency comparisons, we continue to discuss their importance for lane detection, as an empirical recipe provided to readers.

\noindent$\bullet$
\textbf{Task Paradigm.} Segmentation-based methods achieve instance-level discrimination and lane positioning in a two-stage approach. The majority of the algorithm's runtime is occupied by independent instance discrimination processes. This makes them overall less efficient than object detection-based methods which are achieved in one-stage. For object detection-based methods, it is necessary to consider the matching strategy of the positive and negative samples during the network training. This will determine whether NMS is needed for post-processing after the network inference.

\noindent$\bullet$
\textbf{Lane modeling.} In mask-based modeling methods~\cite{SCNN, SAD, RESA}, each pixel is classified, which can lead to inaccurate segmentation masks that subsequently hinder vectorized fitting. Thus, achieving optimal performance and efficiency remains challenging. In contrast, keypoints-based modeling, line anchor-based modeling, and curve-based modeling methods learn fewer points or parameters, directly yielding the vectorized results for downstream use.

Keypoints-based modeling methods~\cite{GANet, CondLSTR} demonstrate strong performance, benefiting from high-precision attitude estimation techniques. However, the overall efficiency of these algorithms is constrained by the instance discrimination step inherent in their segmentation paradigms. 

Line anchor-based modeling methods~\cite{LaneATT, CLRNet, Anchor3DLane, LATR} leverage the vertical and elongated characteristics of lanes in monocular images to strike a good balance between performance and efficiency. Nonetheless, these methods, which learn the horizontal offsets of equidistant points, are unsuitable for U-shaped or nearly horizontal lanes. This corner case is further discussed in subsequent sections.

Curve-based modeling methods~\cite{LSTR, BezierLaneNet} exhibit decent efficiency but fall short in terms of competitive performance on 2D lane detection benchmarks. Interestingly, this kind of method achieves strong results in 3D lane detection~\cite{CLGo, LaneCPP}. As analyzed by Han et al.~\cite{DecoupleLane}, this discrepancy is due to the ground height influence, which makes fitting irregular lanes challenging in FV. In contrast, these lanes appear smooth in BEV, where they can be more easily fitted.

Finally, grids-based modeling methods like UFLD~\cite{UFLD} achieve the highest efficiency; however, this comes at the cost of reduced computational load, resulting in suboptimal performance. These methods often require more advanced operators to compensate for this trade-off~\cite{CondLaneNet}.

\noindent$\bullet$
\textbf{Global context supplementing.} Regardless of the genre, most methods converge on the consensus that supplementing global information significantly enhances lane detection performance, particularly for detecting occluded lanes. Additionally, it is crucial to ensure that these specially designed structures achieve a balance between efficient processing and effective results. While this aspect has received limited attention in existing 3D lane detection benchmarks and methods, in practical applications, certain solutions in 2D lane detection can provide valuable references or be seamlessly integrated into 3D lane detection frameworks.

\noindent$\bullet$
\textbf{Perspective effect elimination.} The ultimate goal persists in obtaining precise 3D lanes to support downstream applications. Using IPM to project 2D lane detection results into 3D space is feasible. However, the assumption of a flat ground often yields incorrect results in BEV, even if predictions are accurate in FV. While projecting the 2D lane detection results into 3D space based on depth values~\cite{ONCE3DLanes} is straightforward, this approach depends heavily on depth estimation and cannot be optimized in an end-to-end manner.

Early 3D lane detection methods~\cite{3DLaneNet, GenLaneNet, CLGo}, which still assume a flat ground, leverage IPM to construct BEV features. Some later approaches~\cite{PersFormer, BEVLaneDet, LaneCPP} improve on this by incorporating learnable ways, leading to enhanced performance. Alternatively, other methods~\cite{CurveFormer, Anchor3DLane, LATR} avoid BEV feature construction entirely, modeling 3D lanes directly and employing a 3D-to-2D forward projection to circumvent the inherent errors introduced by IPM. It should be pointed out that Transformer~\cite{Transformer} has strong abilities in view transformation of BEV features~\cite{PersFormer} or interaction between 3D lanes and FV features~\cite{LATR}. This conclusion is also widely confirmed in related 3D object detection works~\cite{BEVFormer, PETR}. Nonetheless, the hardware deployment of advanced operators, such as deformable attention~\cite{DeformableDETR}, remains a problem worth of optimization.
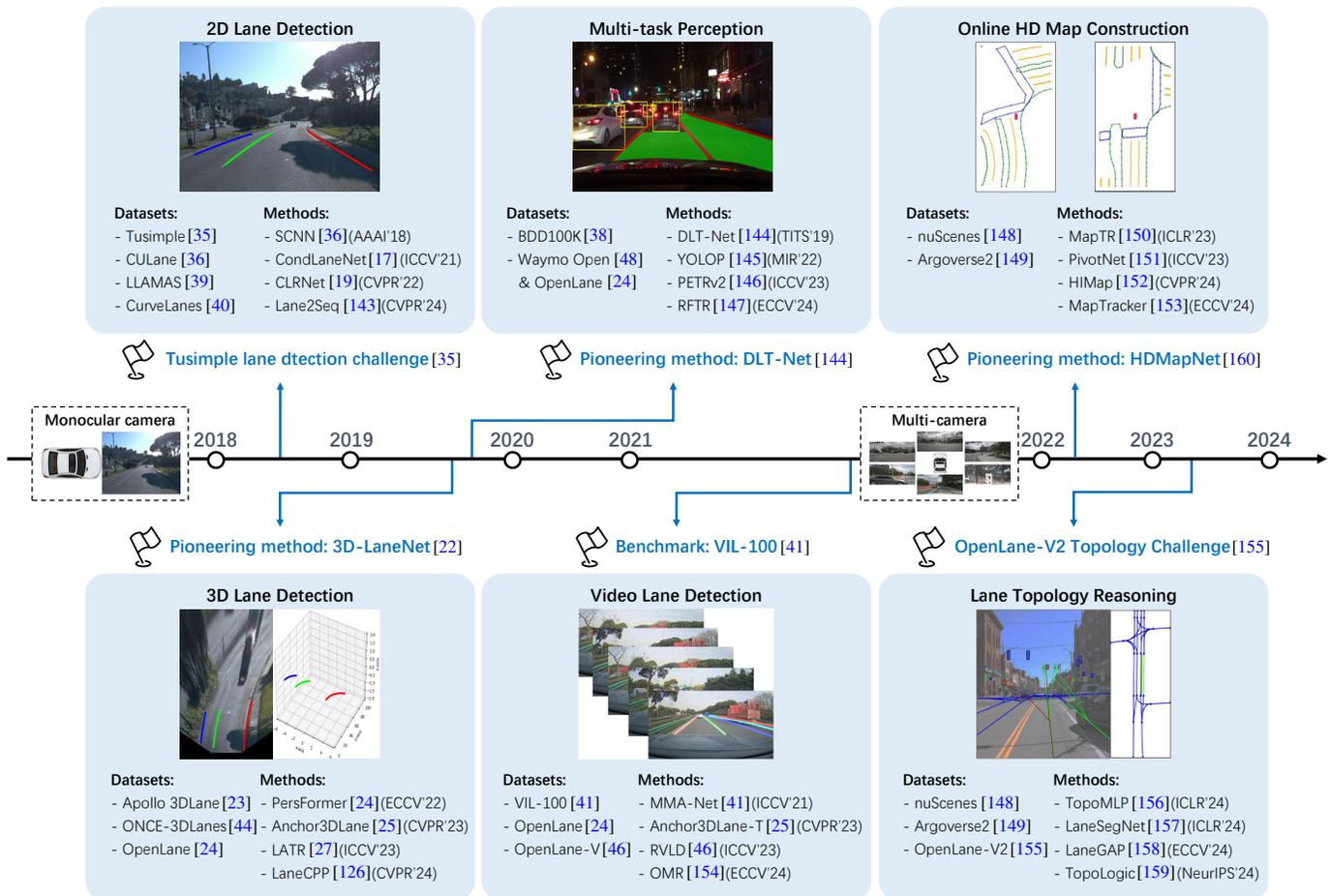
\begin{figure*}
\centering
\begin{overpic}[width=1.0\linewidth]{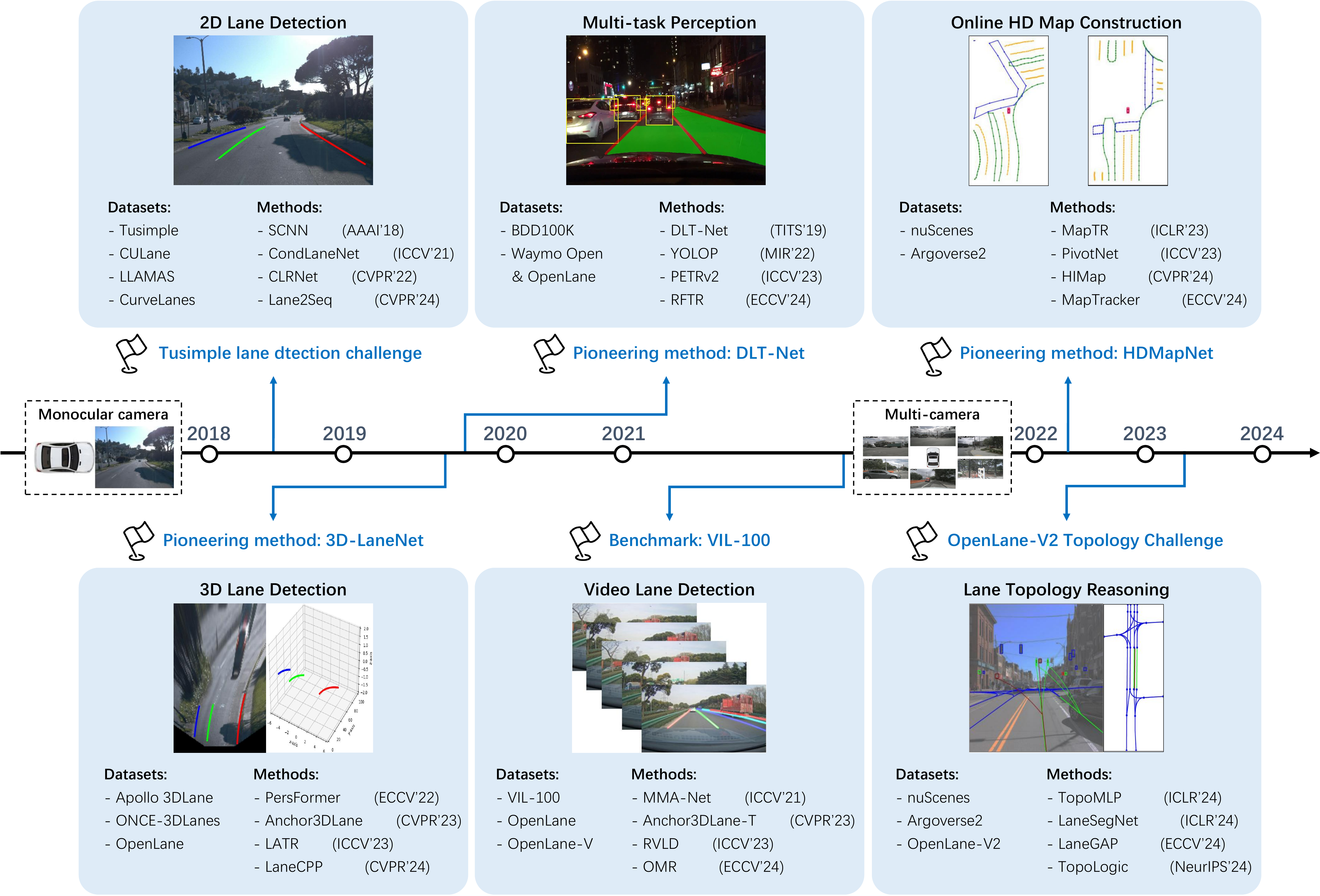}
\put(13.2,50.1){\scriptsize~\cite{Tusimple}}
\put(12.7,48.3){\scriptsize~\cite{SCNN}}
\put(13.0,46.6){\scriptsize~\cite{LLAMAS}}
\put(14.6,44.8){\scriptsize~\cite{CurveLanes}}

\put(23.1,50.1){\scriptsize~\cite{SCNN}}
\put(26.9,48.3){\scriptsize~\cite{CondLaneNet}}
\put(23.9,46.6){\scriptsize~\cite{CLRNet}}
\put(25.0,44.8){\scriptsize~\cite{Lane2Seq}}
\put(43.2,50.1){\scriptsize~\cite{BDD100K}}
\put(45.5,48.3){\scriptsize~\cite{Waymo}}
\put(45.0,46.6){\scriptsize~\cite{PersFormer}}

\put(54.9,50.1){\scriptsize~\cite{DLTNet}}
\put(54.2,48.3){\scriptsize~\cite{YOLOP}}
\put(54.2,46.6){\scriptsize~\cite{PETRv2}}
\put(53.0,44.8){\scriptsize~\cite{RepVF}}
\put(73.5,50.1){\scriptsize~\cite{nuScenes}}
\put(74.4,48.3){\scriptsize~\cite{Argoverse2}}

\put(83.7,50.1){\scriptsize~\cite{MapTR}}
\put(84.5,48.3){\scriptsize~\cite{PivotNet}}
\put(83.5,46.6){\scriptsize~\cite{HIMap}}
\put(86.1,44.8){\scriptsize~\cite{MapTracker}}
\put(15.8,7.1){\scriptsize~\cite{GenLaneNet}}
\put(16.3,5.3){\scriptsize~\cite{ONCE3DLanes}}
\put(13.7,3.6){\scriptsize~\cite{PersFormer}}

\put(25.5,7.1){\scriptsize~\cite{PersFormer}}
\put(27.2,5.3){\scriptsize~\cite{Anchor3DLane}}
\put(22.4,3.6){\scriptsize~\cite{LATR}}
\put(24.2,1.8){\scriptsize~\cite{LaneCPP}}
\put(42.2,7.1){\scriptsize~\cite{VIL100}}
\put(43.3,5.3){\scriptsize~\cite{PersFormer}}
\put(44.6,3.6){\scriptsize~\cite{RVLD}}

\put(53.6,7.1){\scriptsize~\cite{VIL100}}
\put(57.0,5.3){\scriptsize~\cite{Anchor3DLane}}
\put(51.1,3.6){\scriptsize~\cite{RVLD}}
\put(51.0,1.8){\scriptsize~\cite{OMR}}
\put(73.3,7.1){\scriptsize~\cite{nuScenes}}
\put(74.2,5.3){\scriptsize~\cite{Argoverse2}}
\put(75.5,3.6){\scriptsize~\cite{OpenLane-V2}}

\put(84.7,7.1){\scriptsize~\cite{TopoMLP}}
\put(85.9,5.3){\scriptsize~\cite{LaneSegNet}}
\put(84.4,3.6){\scriptsize~\cite{LaneGAP}}
\put(85.1,1.8){\scriptsize~\cite{Topologic}}
\put(31.7,40.7){\scriptsize~\cite{Tusimple}}
\put(60.7,40.7){\scriptsize~\cite{DLTNet}}
\put(91.6,40.7){\scriptsize~\cite{HDMapNet}}

\put(31.8,26.5){\scriptsize~\cite{3DLaneNet}}
\put(58.1,26.5){\scriptsize~\cite{VIL100}}
\put(92.3,26.5){\scriptsize~\cite{OpenLane-V2}}
\end{overpic}
\caption{
The roadmap for the development of hotspots in lane detection research. In each domain, a symbolic event is identified as a milestone and presented in chronological order. The representative datasets and methods within each task are listed.
}
\label{fig:roadmap}
\vspace{-0.4cm}
\end{figure*}

\section{Extended Works of Lane Detection}
\label{sec:extended_work}
There are also some works that have received widespread attention in recent years, which are closely related to lane detection. In terms of task flow, they can be regarded as an upgrade on monocular image lane detection. We provide a brief introduction to them in this section. Figure~\ref{fig:roadmap} depicts a roadmap of the evolution from lane detection to its expansion works.

\subsection{Multi-task Perception}
\label{sec:multi-task}
In autonomous driving, multiple perception tasks often need to be processed synchronously, in real-time, and in parallel. A shared backbone can save computation costs and improve efficiency greatly. Thus, leveraging a unified framework to conduct multiple perception tasks simultaneously gradually becomes a research hotspot. Early works~\cite{DLTNet, YOLOP, HybridNets, YOLOPv2, TwinLiteNet, AYOLOM} connect multiple specific task heads after the feature extractor to simultaneously complete three tasks on the BDD100K~\cite{BDD100K} dataset: object detection, drivable area segmentation, and lane detection. These methods achieve impressive results in each task, which benefit from the powerful and efficient encoder and carefully designed multi-task learning strategy. However, the labels of lanes in BDD100K are only semantic-level annotations, and only binary segmentation methods can be used. Further post-processing is needed to distinguish each lane instance. Recent researches mainly focus on multi-task 3D perception. PETRv2~\cite{PETRv2} designs detection query, segmentation query, and lane query to support 3D object detection, BEV segmentation, and 3D lane detection simultaneously. Li et al.~\cite{RepVF} propose a unified representation method for multiple perception tasks. They represent 3D objects and 3D lanes as a kind of 3D vector field, which allows them to leverage a single-head unified model to achieve multi-task perception.

\subsection{Video Lane Detection}
\label{sec:temporal}
As mentioned in the previous chapters, current works attempt to supplement more global information to better detect lanes with unclear visual clues. However, these methods rely on detectors that use single images. In autonomous driving systems, video frames are captured continuously. Therefore, the correlation between frames can be used to more reliably detect obscure lanes in the current frame. For 2D lane detection, Zou et al.~\cite{RobustLD} and Zhang et al.~\cite{LDspatio} use recursive neural networks to fuse the features of the current frame with those of several past frames. Zhang et al.~\cite{VIL100} aggregate the features of the current frame and multiple past frames based on Transformer. Tabelini et al.~\cite{LDS2022} extract lane features from video frames using LaneATT~\cite{LaneATT} and combine these features. Wang et al.~\cite{VLD2022} utilize spatiotemporal information from adjacent video frames by extending the feature aggregation module in RESA~\cite{RESA}. Jin et al. put RVLD~\cite{RVLD}, which includes an intra-frame lane detector to locate lanes in stationary frames and a predictive lane detector to use information from the previous frame for lane detection in the current frame. OMR~\cite{OMR} employs vehicle masks occupying lanes to interact with historical frames, further improving the accuracy of lane detection in the current frame. For 3D lane detection, STLane3D~\cite{STLane3D} proposes a multi-frame pre-alignment layer under the BEV space, which uniformly projects features from different frames onto the same ROI region. Anchor3DLane-T~\cite{Anchor3DLane} incorporates temporal information by projecting the 3D anchors of the current frame onto previous frames to sample features. CurveFormer++~\cite{CurveFormer++} designs a temporary Curve Cross Attention module based on CurveFormer~\cite{CurveFormer}, which can selectively utilize historical curve query and keypoints to propagate historical information frame by frame.

\subsection{Online HD Map Construction}
\label{sec:hdmap}
HD maps are an essential module for autonomous driving. Although the traditional offline method of building HD maps can generate accurate map information and is adopted by many autonomous driving companies, it requires a lot of manual annotation costs. As an alternative, an increasing number of works try to design a novel HD map learning framework that makes use of on-car sensors and computation to estimate vectorized local semantic maps.

From a process perspective, they typically follow the general pipeline of BEV perception tasks~\cite{BEVsurvey}, taking multi-camera images as input, extracting image features using a 2D encoder, then obtaining BEV features through a view transform module, and finally outputting various map elements from the BEV perspective through a specific map element decoder. Due to the increase in the number of sensors and the fact that the map elements to be detected include but are not limited to lanes, pedestrian crossings, lane separations, and lane boundaries, the task is more challenging than monocular lane detection. Similarly, the key to this type of work is how to model map elements with different shapes, such as lines and polygons, into a set of values that can be learned through neural networks.

HDMapNet~\cite{HDMapNet} adopts a basic bottom-up segmentation approach to perform semantic segmentation on all map elements. Then it combines instance embedding and post-processing clustering to obtain each map element instance. This rasterization result still requires post-processing for downstream use, therefore, subsequent work attempts to predict vectorized maps end-to-end. BeMapNet~\cite{BeMapNet} models map elements as segmented B\'{e}zier curves. It detects map elements first and then regresses the detailed points with a piecewise B\'{e}zier head. VectorMapNet~\cite{VectorMapNet} uses a line to represent all map elements and defines a hierarchical query representation. The points of the map elements are autoregressively output through the transformer decoder. However, it outputs the point set through autoregression, which leads to low efficiency. To solve this problem,  MapTR~\cite{MapTR} designs a unique representation method for map elements. It uses lines and polygons with uniform sampling points to represent line and area elements, respectively. Therefore, all map elements are represented as sets with the same number of points and different arrangement orders. Owing to its unified permutation-equivalent modeling approach and hierarchical query design, MapTR achieves advanced performance and efficiency on the nuScenes~\cite{nuScenes} dataset solely with camera input, providing a solid baseline for follow-up research. Afterward, MapTRv2~\cite{MapTRv2} improves self-attention and cross-attention in the decoder of MapTR, further enhancing both the accuracy and performance. PivotNet~\cite{PivotNet} proposes an end-to-end framework for representing map elements using pivot points. The purpose is to address the issue of shape information loss caused by using a fixed number and consistent position of points to represent complex map elements in MapTR. HIMap~\cite{HIMap} meticulously designs feature extractors for MapTR's hierarchical query, enabling the model to better learn instance-level features. StreamMapNet~\cite{StreamMapNet} improves MapTR in terms of timing. It overlays information from all historical frames together and implements a memory mechanism using recurrent late embedding. Then MapTracker~\cite{MapTracker} formalizes the online HD map construction as a tracking task and uses the history of memory latents to ensure consistency in reconstruction over time.

\subsection{Lane Topology Reasoning}
\label{sec:topology}
Topology reasoning aims to comprehensively understand road scenes and present drivable routes in autonomous driving. It requires detecting road centerlines and traffic elements, further reasoning their topology relationship, i.e., lane-lane topology, and lane-traffic topology. Directly using vehicle-mounted sensors to detect lane topology has become popular due to their practical value.

The early topology reasoning works mainly focus on lane-lane topology, i.e., detecting the centerlines to construct a lane graph. Extraction of the lane topology task is first proposed by STSU~\cite{STSU}, which predicts the centreline and lane connectivity relationships. TopoRoad~\cite{TopoRoad} uses a set of directed lane curves and their interactions to represent road topology. Can et al.~\cite{Can2023online} provide additional supervision of the relationship by considering the centerlines as cluster centers to assign objects. LaneGAP~\cite{LaneGAP} utilizes the shortest path algorithm in graph theory to transform lane topology into a series of overlapping paths and directly obtains information about these complete paths through end-to-end learning. CenterLineDet~\cite{CenterLineDet} regards centerlines as vertices and designs a graph model to update centerline topology. Recently, lane-traffic topology is additionally introduced by the OpenLane-V2~\cite{OpenLane-V2} dataset, further improving the understanding of scene structure. Aiming at a complete and diverse driving scene graph, TopoNet~\cite{TopoNet} explicitly models the connectivity of centerlines within the network and incorporates traffic elements into the task. TopoMLP~\cite{TopoMLP} leverages position embedding~\cite{PETR} to enhance topology modeling. LaneSegNet~\cite{LaneSegNet} proposes a unified representation for integrating lanes and centerlines. It introduces a lane attention mechanism to facilitate the learning of topological relationships between centerlines and lanes. To improve lane topology inference, TopoLogic~\cite{Topologic} introduces an efficient post-processing that integrates the geometric distance between centerline endpoints and the semantic similarity of lane queries within a high-dimensional space.

\section{Future Direction}
\label{sec:future_direction}
This section outlines potential directions for future research in lane detection. The scope of our discussion includes the improvable issues within the field, underexplored subfields, and relevant tasks outside this area that hold significant research value.

\noindent$\bullet$
\textbf{General and Unified Lane Modeling.}
Effectively modeling lanes of arbitrary shapes without compromising efficiency remains a significant challenge. When the scenario extends from a monocular camera's front view to multi-camera surround views, the presence of numerous U-shaped or nearly horizontal lanes becomes common. In such cases, modeling approaches that heavily rely on prior knowledge, such as grids-based modeling for row-wise classification, or line anchor-based modeling for learning the longitudinal equidistant offset points, are unsuitable. In contrast, mask-based modeling methods are more reliable despite their lower performance and efficiency. The existing vectorized map element modeling methods~\cite{VectorMapNet, MapTR} provide valuable guidance on this issue. Recently, Lane2Seq~\cite{Lane2Seq} unifies 2D lane detection through sequence generation. This approach also serves as a promising direction for subsequent studies, although its efficiency still requires further improvement.

\noindent$\bullet$
\textbf{Multi-modal Lane Detection.} 
In recent years, LiDAR-based lane detection benchmarks and methods~\cite{kammel2008lidar, thuy2010lane, jung2018real, paek2022row, paek2022k, liu2023learning, guan2023flexible} gain attention as another minority approach. Although the 3D information can be directly offered by LiDAR, its shorter perception range and high cost have made camera-based methods more prevalent. However, LiDAR provides the advantages by remaining unaffected by lighting changes and delivering accurate depth information, effectively compensating for the limitations of cameras. The integration of LiDAR and camera data demonstrates significant effectiveness in enhancing performance, which is widely validated in the domain of 3D perception~\cite{liu2023bevfusion, liang2022bevfusion, UniFusion, UniTR}. However, there are notably few dedicated works focusing on multi-modal lane detection~\cite{bai2018deep, M23DLaneNet, DV3DLane}.

\noindent$\bullet$
\textbf{Label Efficient Lane Detection.} 
The existing lane detection methods mainly focus on supervised learning, which needs a lot of annotations for training, which leads to huge manual costs. Thus, developing annotation-efficient lane detection algorithms is necessary. WS-3D-Lane~\cite{WS3DLane} uses 2D lane labels to weakly supervise 3D lane detection, which is valuable for research and mass production. Furthermore, unsupervised lane detection~\cite{CLLD, LDMAE, LaneCorrect, MLDA} is also a promising direction, although the related works are limited.

\noindent$\bullet$
\textbf{Lane Detection in End-to-End Autonomous Driving.}
The CVPR Best Paper, UniAD~\cite{UniAD}, attracts significant interest~\cite{FusionAD, VAD, GenAD, SparseDrive, VADv2} in both academia and industry regarding the development of end-to-end autonomous driving systems. Unlike the conventional modular architecture of 'perception-prediction-planning', end-to-end autonomous driving directly outputs vehicle motion planning results from sensor data in a fully differentiable manner. Within this framework, lane detection no longer outputs explicit lane coordinate values but instead functions as a module providing intermediate representations of lanes. However, this approach often encounters challenges such as limited interpretability and inadequate generalization, particularly in complex road scenarios, including curved roads and multi-lane switching. Future research might explore hybrid architectures that incorporate specific lane detection outputs, such as lane centerlines, lane width, and curvature, as prior knowledge into intermediate representations within end-to-end models. This integration can enable the network to better capture the structural characteristics of roads and ensure the preservation of critical lane-level information in the decision-making process.

\noindent$\bullet$
\textbf{Visual Reasoning for Lane Detection.} 
The advent of large language models (LLMs) and vision-language Models (VLMs) unlock significant potential for multimodal artificial intelligence systems to perceive the real world, make decisions, and control tools with a capability akin to human cognition. Recently, LLMs and VLMs are integrated into autonomous driving systems, primarily focusing on visual reasoning related to dynamic objects, the generation of future trajectories, and the detailed control signals of ego-vehicles~\cite{DiLu, DriveGPT4, DriveLM}. In contrast, relatively few studies explore visual inference concerning static objects, such as lanes. Fortunately, a new benchmark~\cite{MapLM} specifically designed for large-scale visual reasoning in understanding maps and traffic scenarios is emerged. By training on extensive traffic scene data, the models can derive insights from complex multi-modal driving resources, including map data, traffic regulations, and incident reports. This enables them to enhance vehicle navigation and planning with safety and efficiency parameters, while also adapting to dynamic road conditions with an understanding that closely resembles human intuition. 

\noindent$\bullet$
\textbf{Roadside Lane Detection.} 
The current perception capabilities in autonomous driving primarily focus on ego vehicles. While vehicle-based perception systems capture the immediate surrounding environment, their range is limited to short distances. In contrast, roadside cameras, mounted on utility poles several meters above ground, enable remote perception with minimal visual obstructions. Recently, roadside 3D object detection datasets~\cite{Rope3d, Dair-v2x} and corresponding methods~\cite{BEVHeight, MonoUNI, CoBEV} are developed to promote 3D perception tasks in roadside scenes, facilitating potential collaboration between vehicles and infrastructure. However, there are still no established benchmarks or methods specifically for roadside lane detection. Roadside lane detection can effectively substitute for manual monitoring of lane violations or illegal lane changes, offering significant potential for applications in security.

\section{Conclusion}
\label{sec:conclusion}
This survey comprehensively reviews the latest progress in monocular lane detection based on deep learning, covering both 2D and 3D lane detection methods in recent years.
Four core designs in lane detection algorithms are identified through theoretical analysis and experimental evaluation: (1) Task paradigm, focusing on lane instance-level discrimination; (2) Lane modeling, representing lanes as a set of learnable parameters in the neural network; (3) Global information supplementation, enhancing the inference on the obscure lanes; (4) Perspective effect elimination, providing accurate 3D lanes for downstream applications. From these perspectives, this paper presents a comprehensive overview of existing methods.
In addition, this article also reviews extended works on monocular lane detection to provide readers with a more comprehensive understanding of the development of lane detection.
Finally, the future research directions for lane detection are pointed out.

\appendix
\noindent
\textbf{Overview.} In this appendix, we provide more details as a supplementary adjunct to the main paper.

\begin{enumerate}
\setlength{\leftmargin}{-1em}
\setlength{\parsep}{0ex} 
\setlength{\topsep}{0ex}
\setlength{\itemsep}{0.5ex}  
\setlength{\labelsep}{0.5em} 
\setlength{\itemindent}{-0.5em} 
\setlength{\listparindent}{0em} 
\item More descriptions on task metrics. (Section~\ref{sec:more_metric})
\item More benchmark results. (Section~\ref{sec:more_benchmark_results})
\item Correspondence between 3D lanes and images. (Section~\ref{sec:camera_model})
\end{enumerate}


\subsection{More Task Metrics}
\label{sec:more_metric}
In this section, we present detailed descriptions of more indicators for lane detection task metrics. 

\noindent$\bullet$
\textbf{Accuracy (Acc).} For Tusimple~\cite{Tusimple} dataset, accuracy will also be used as an indicator, and the evaluation formula is
\begin{equation}\label{acc}
Accuracy = \frac{\sum_{clip}C_{clip}}{\sum_{clip}S_{clip}},
\end{equation}
where $C_{clip}, S_{clip}$ are the number of correct points and the number of ground truth points of an image respectively.

\noindent$\bullet$
\textbf{Average Precision (AP).} It is more often used to evaluate Apollo 3DLane~\cite{GenLaneNet}. As described in Section~\ref{sec:metric}, the TP under different thresholds can be obtained by selecting the decision criteria for TP and iterating the lane confidence thresholds. Then the exact recall curve can be generated and the AP can be obtained by calculating the area under this curve.

\noindent$\bullet$
\textbf{X Error and Z Error in 3D Lane Detection.} When GT matches the corresponding predicted lane, $x/z$ $error$ is defined as
\begin{equation}
X_{Error} = \frac{1}{N} \sum_{i=1}^{N} \sqrt{(x_i - \hat{x_i})^2},
\end{equation}
\begin{equation}
Z_{Error} = \frac{1}{N} \sum_{i=1}^{N} \sqrt{(z_i - \hat{z_i})^2},
\end{equation}
Where $x_i/z_i$ is the $x/z$ coordinate of the GT sampling point, $\hat{x_i}/\hat{z_i}$ is the $x/z$ coordinate of the matched prediction point, and $N$ is the number of points on the lane.

\noindent$\bullet$
\textbf{Chamfer Distance (CD).} This metric proposed by ONCE-3DLanes~\cite{ONCE3DLanes} is used to calculate the curve matching error in the camera coordinate system. The curve matching error $CD_{p,g}$ between $L^{p}$ and $L^{p}$ is calculated as follows:
\begin{small}
\begin{equation}
\left\{
\begin{aligned}
& CD_{p,g} = \frac{1}{m} \sum_{i=1}^{m} || P_{g_i} - \hat{P}_{p_j}||_2, \\
&\hat{P}_{p_j} = \mathop{min}\limits_{P_{p_j}\in L^p}||P_{p_j}-P_{g_i}||_2, \\
\end{aligned}
\right.
\end{equation}
\end{small}
where $P_{p_j}=(x_{p_j}, y_{p_j}, z_{p_j})$ and $P_{g_i}=(x_{g_i}, y_{g_i}, z_{g_i})$ are point of $L^p$ and $L^g$ respectively, and $\hat{P}_{p_j}$ is the nearest point to the specific point $P_{g_i}$.
$m$ represents the number of points token at an equal distance from the ground-truth lane.

\subsection{More Benchmark Results}
\label{sec:more_benchmark_results}
\noindent$\bullet$
\textbf{Results on other 2D lane detection datasets.} Table~\ref{tab:other_2d_datasets_results} shows the performance comparison on Tusimple~\cite{Tusimple}, LLAMAS~\cite{LLAMAS} and CurveLanes~\cite{CurveLanes}.

\begin{table}[ht]
\caption{Benchmark results of representative 2D lane detection methods on Tusimple~\cite{Tusimple}, LLAMAS~\cite{LLAMAS} and CurveLanes~\cite{CurveLanes}.}
\centering
\resizebox{0.48\textwidth}{!}{\begin{tabular}{lccccc} 
\toprule
\multirow{2}{*}{\textbf{Method}} & \multirow{2}{*}{\textbf{Backbone}} & \multicolumn{2}{c}{\textbf{Tusimple}} & \textbf{LLAMAS-Test} & \textbf{CurveLanes} \\ 
\cmidrule(lr){3-4} \cmidrule(lr){5-5} \cmidrule(lr){6-6} 
 & & \textbf{F1($\%$)}& \textbf{Acc($\%$)} & \textbf{F1($\%$)} & \textbf{F1($\%$)} \\
\midrule
\multicolumn{6}{l}{Segmentation based-methods} \\
\midrule
LaneNet~\cite{LaneNet} & ENet & - & 96.40 & - & - \\
SCNN~\cite{SCNN} & VGG16 & 95.97 & 96.53 & 95.16 & - \\
SAD~\cite{SAD} & ENet & 95.97 & 96.53 & - & 50.31 \\
RESA~\cite{RESA}  & ResNet34 & 96.93 & 96.82 & - & - \\
LaneAF~\cite{LaneAF} & DLA34 & 96.49 & 95.62 & 96.90 & - \\
CondLaneNet~\cite{CondLaneNet} & ResNet18 & 97.01 & 95.48 & - & 85.09 \\
UFLDv2~\cite{UFLDv2} & ResNet18 & 96.05 & 95.50 & 94.58 & 80.45 \\
FOLOLane~\cite{FOLOLane} & ERFNet & 96.59 & 96.92 & - & - \\
GANet~\cite{GANet} & ResNet18 & 97.71 & 95.95 & - & - \\
RCLane~\cite{RCLane} & ResNet18 & 97.52 & 96.49 & 96.05 & 90.47 \\ 
CondLSTR~\cite{CondLSTR} & ResNet18 & 97.71 & 96.06 & - & 87.99 \\ 
\midrule
\multicolumn{6}{l}{Object detection based-methods} \\
\midrule
CurveLane-S~\cite{CurveLanes} & - & - & - & 90.2 & 81.12 \\
PolyLaneNet~\cite{PolyLaneNet} & EfficientNetB0 & - & 93.36 & 90.2 & - \\
LaneATT~\cite{LaneATT} & ResNet18 & 96.71 & 95.57 & 93.46 & - \\
LaneAF~\cite{LaneAF} & DLA34 & 96.49 & 95.62 & 96.90 & - \\ 
BézierLaneNet~\cite{BezierLaneNet} & ResNet18 & - & 95.41 & 94.91 & - \\ 
CLRNet~\cite{CLRNet} & ResNet18 & 97.89 & 96.84 & 96.00 & - \\
PGA-Net~\cite{PGANet} & ResNet18 & 97.66 & 95.43 & 94.18 & - \\
\bottomrule
\end{tabular}}
\label{tab:other_2d_datasets_results}
\end{table}

\noindent$\bullet$
\textbf{Results on other 3D lane detection datasets.} We report the performance comparison on ONCE-3DLanes~\cite{ONCE3DLanes} in Table~\ref{tab:once3dlane_result}, and Apollo 3DLane~\cite{GenLaneNet} in Table~\ref{tab:apollo_result}.

\begin{table}[ht]
\centering
\caption{Benchmark results of representative 3D lane detection methods on Apollo 3DLane~\cite{GenLaneNet}.}
\resizebox{0.48\textwidth}{!}{\begin{tabular}{c|lccccccc}
\toprule
\multirow{2}{*}{\textbf{Scene}} & \multirow{2}{*}{\textbf{Method}} & \multirow{2}{*}{\textbf{Backbone}} & \multirow{2}{*}{\textbf{AP($\%$)}} & \multirow{2}{*}{\textbf{F1($\%$)}} & \multicolumn{2}{c}{\textbf{X error(m)}} & \multicolumn{2}{c}{\textbf{Z error(m)}} \\ 
\cmidrule(lr){6-7} \cmidrule(lr){8-9} 
 & & & & & \textbf{near} & \textbf{far} & \textbf{near} & \textbf{far} \\
\midrule
\multirow{9}{*}{\makecell[c]{Balanced \\ Scene} }
 & 3DLaneNet~\cite{3DLaneNet} & VGG16 & 89.3 & 86.4 & 0.068 & 0.477 & 0.015 & 0.202 \\ 
 & Gen-LaneNet~\cite{GenLaneNet} & ERFNet & 90.1 & 88.1 & 0.061 & 0.496 & 0.012 & 0.214 \\ 
 & CLGo~\cite{CLGo} & VGG16 & 94.2 & 91.9 & 0.061 & 0.361 & 0.029 & 0.250 \\ 
 & PersFormer~\cite{PersFormer} & EfficientNetB7 & - & 92.9 & 0.054 & 0.356 & 0.010 & 0.234 \\ 
 & Reconstruct~\cite{li2022reconstruct} & ERFNet & 93.8 & 91.9 & 0.049 & 0.387 & 0.008 & 0.213 \\ 
 & CurveFormer~\cite{CurveFormer} & EfficientNetB7 & 97.3 & 95.8 & 0.078 & 0.326 & 0.018 & 0.219 \\ 
 & Anchor3DLane~\cite{Anchor3DLane} & ResNet18 & 97.2 & 95.6 & 0.052 & 0.306 & 0.015 & 0.223 \\ 
 & BEV-LaneDet~\cite{BEVLaneDet} & ResNet34 & - & 98.7 & 0.016 & 0.242 & 0.020 & 0.216 \\ 
 & LATR~\cite{LATR} & ResNet50 & 97.9 & 96.8 & 0.022 & 0.253 & 0.007 & 0.202 \\ 
 & LaneCPP~\cite{LaneCPP} & EfficientNetB7 & 99.5 & 97.4 & 0.030 & 0.277 & 0.011 & 0.206 \\ 
\midrule
\multirow{9}{*}{\makecell[c]{Rarely \\ Observed}} 
 & 3DLaneNet~\cite{3DLaneNet} & VGG16 & 74.6 & 72.0 & 0.166 & 0.855 & 0.039 & 0.521 \\ 
 & Gen-LaneNet~\cite{GenLaneNet} & ERFNet & 79.0 & 78.0 & 0.139 & 0.903 & 0.030 & 0.539 \\ 
 & CLGo~\cite{CLGo} & VGG16 & 88.3 & 86.1 & 0.147 & 0.735 & 0.071 & 0.609 \\ 
 & PersFormer~\cite{PersFormer} & EfficientNetB7 & - & 87.5 & 0.107 & 0.782 & 0.024 & 0.602 \\ 
 & Reconstruct~\cite{li2022reconstruct} & ERFNet & 85.2 & 83.7 & 0.126 & 0.903 & 0.023 & 0.625 \\ 
 & CurveFormer~\cite{CurveFormer} & EfficientNetB7 & 97.1 & 95.6 & 0.182 & 0.737 & 0.039 & 0.561 \\ 
 & Anchor3DLane~\cite{Anchor3DLane} & ResNet18 & 96.9 & 94.4 & 0.094 & 0.693 & 0.027 & 0.579 \\ 
 & BEV-LaneDet~\cite{BEVLaneDet} & ResNet34 & - & 99.1 & 0.031 & 0.594 & 0.040 & 0.556 \\ 
 & LATR~\cite{LATR} & ResNet50 & 97.3 & 96.1 & 0.050 & 0.600 & 0.015 & 0.532 \\ 
 & LaneCPP~\cite{LaneCPP} & EfficientNetB7 & 98.6 & 96.2 & 0.073 & 0.651 & 0.023 & 0.543 \\  
\midrule
\multirow{9}{*}{\makecell[c]{Visual \\ Variations}} 
 & 3D-LaneNet~\cite{3DLaneNet} & VGG16 & 74.9 & 72.5 & 0.115 & 0.601 & 0.032 & 0.230 \\ 
 & Gen-LaneNet~\cite{GenLaneNet} & ERFNet & 87.2 & 85.3 & 0.074 & 0.538 & 0.015 & 0.232 \\ 
 & CLGo~\cite{CLGo} & VGG16 & 89.2 & 87.3 & 0.084 & 0.464 & 0.045 & 0.312 \\ 
 & PersFormer~\cite{PersFormer} & EfficientNetB7 & - & 89.6 & 0.074 & 0.430 & 0.015 & 0.266 \\ 
 & Reconstruct~\cite{li2022reconstruct} & ERFNet & 92.1 & 89.9 & 0.060 & 0.446 & 0.011 & 0.235 \\ 
 & CurveFormer~\cite{CurveFormer} & EfficientNetB7 & 93.0 & 90.8 & 0.125 & 0.410 & 0.028 & 0.254 \\ 
 & Anchor3DLane~\cite{Anchor3DLane} & ResNet18 & 93.6 & 91.4 & 0.068 & 0.367 & 0.020 & 0.232 \\ 
 & BEV-LaneDet~\cite{BEVLaneDet} & ResNet34 & - & 96.9 & 0.027 & 0.320 & 0.031 & 0.256 \\ 
 & LATR~\cite{LATR} & ResNet50 & 96.6 & 95.1 & 0.045 & 0.315 & 0.016 & 0.228 \\ 
 & LaneCPP~\cite{LaneCPP} & EfficientNetB7 & 93.7 & 90.4 & 0.054 & 0.327 & 0.020 & 0.222 \\ 
\bottomrule
\end{tabular}}
\label{tab:apollo_result}
\end{table}

\begin{table}[ht]
\centering
\caption{Benchmark results of representative 3D lane detection methods on ONCE-3DLanes~\cite{ONCE3DLanes}.}
\resizebox{0.48\textwidth}{!}{\begin{tabular}{lccccc}
\midrule
\textbf{Method} & \textbf{Backbone} & \textbf{F1($\%$)} & \textbf{Precision($\%$)} & \textbf{Recall($\%$)} & \textbf{CD Error(m)} \\ 
\midrule
3D-LaneNet~\cite{3DLaneNet} & VGG16 & 44.73 & 61.46 & 35.16 & 0.127 \\ 
Gen-LaneNe~\cite{GenLaneNet} & ERFNet & 45.59 & 63.95 & 35.42 & 0.121 \\ 
SALAD~\cite{ONCE3DLanes} & SegFormer & 64.07 & 75.90 & 55.42 & 0.098 \\ 
PersFormer~\cite{PersFormer} & EfficientNetB7 & 74.33 & 80.30 & 69.18 & 0.074 \\ 
Anchor3DLane~\cite{Anchor3DLane} & ResNet18 & 74.87 & 80.85 & 69.71 & 0.060 \\
LATR~\cite{LATR} & ResNet50 & 80.59 & 86.12 & 75.73 & 0.052 \\
PVALane~\cite{PVALane} & ResNet50 & 76.35 & 70.83 & 82.81 & 0.059 \\
\bottomrule
\end{tabular}
\label{tab:once3dlane_result}}
\end{table}

\subsection{Correspondence between 3D Lanes and Images.}
\label{sec:camera_model}

\begin{figure}[ht]
\begin{center}
\includegraphics[width=0.45\textwidth]{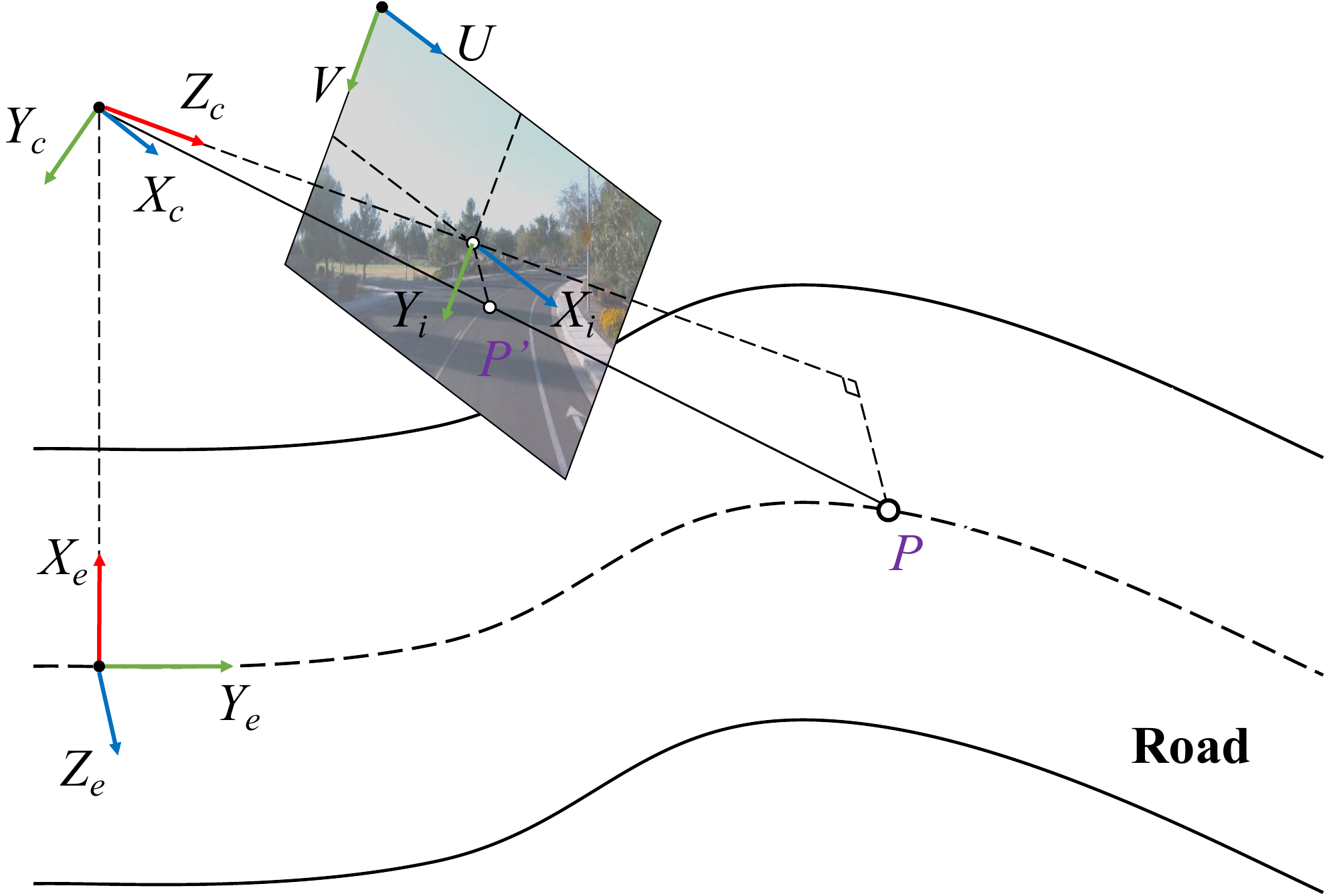}
\end{center}
\caption{Geometry setup about camera and 3D lane. The $(X_e, Y_e, Z_e)$, $(X_c, Y_c, Z_c)$ represent ego-vehicle coordinate and camera coordinate, and the $(X_i, Y_i)$, $(U, V)$ represent image coordinate and pixel coordinate. $P$, $P'$ corresponds to the 3D point from the lane and the projected 2D point from the camera view, respectively. 
Given the ego-vehicle coordinates of $P$ and 
the intrinsic and extrinsic parameters of the camera, the pixel coordinates of $P'$ can be obtained.
}
\label{fig:camera_model}
\end{figure}

This section introduces the correspondence between 3D lanes and images. Figure~\ref{fig:camera_model} depicts the imaging process of 3D lanes in the camera. Utilizing the commonly employed pinhole camera projection as an illustration, the projection process encompasses transformation between the ego-vehicle, camera, image, and pixels.

The transformation from the ego-vehicle coordinate system to the camera coordinate system involves translation and rotation exclusively. Consider $P_e=[x_e,y_e,z_e,1]$, $P_c=[x_c,y_c,z_c,1]$ as the homogenous coordinates of a 3D point $P$ in the ego-vehicle and camera coordinate systems, respectively. Their relationship is elucidated as follows:
\begin{equation}\label{prel-3dv-E1}
P_c = \begin{bmatrix} x_c \\ y_c \\ z_c \\ 1 \end{bmatrix} = 
\begin{bmatrix} \bm{R} & \bm{T} \\ \bm{0}^T & 1 \end{bmatrix}
\begin{bmatrix} x_e \\ y_e \\ z_e \\ 1 \end{bmatrix},
\end{equation}
where $R$, $T$ refer to a rotation matrix and a translation matrix respectively. 

The image coordinate system is employed to represent the perspective projection from the camera coordinate system onto the image plane. When the camera distortion is disregarded, the relationship between a 3D point and its image plane projection can be simplified using a pinhole model. The image coordinates $(x_i, y_i)$ are determined by Eqn.~\ref{prel-3dv-E2}:
\begin{equation}\label{prel-3dv-E2}
\begin{cases}
\begin{aligned}
x_i &= f \cdot \frac{x_c}{z_c} \\
y_i &= f \cdot \frac{y_c}{z_c}
\end{aligned},
\end{cases}
\end{equation}
where $f$ represents the focal length of the camera.

The translation and scaling transformation links the image coordinate framework with the pixel coordinate framework. 
Let $\alpha$ and $\beta$ denote the scaling factors for the x-axis and y-axis, respectively, while $c_x$ and $c_y$ represent the translation values shifting the origin of the coordinate system. The pixel coordinates $(u, v)$ can be mathematically formulated as shown in Eqn.~\ref{prel-3dv-E3}:
\begin{equation}\label{prel-3dv-E3}
\begin{cases}
\begin{aligned}
u &= \alpha x + C_x \\
v &= \beta y + C_y
\end{aligned}.
\end{cases}
\end{equation}

With Eqn.~\ref{prel-3dv-E2} and Eqn.~\ref{prel-3dv-E3}, setting $f_x = \alpha f$, $f_y = \beta f$, we could derive Eqn.~\ref{prel-3dv-E4}:
\begin{equation}\label{prel-3dv-E4}
z_c \begin{bmatrix}
u \\
v \\
1
\end{bmatrix}
= \begin{bmatrix}
f_x & 0 & c_x \\
0 & f_y & c_y \\
0 & 0 & 1
\end{bmatrix}
\begin{bmatrix} x_c \\ y_c \\ z_c \\ \end{bmatrix}.
\end{equation}

To sum up, the relationship between the 3D point $P$ in the ego-vehicle coordinate system and its corresponding projection $P'$ in the pixel coordinate system can be described as:
\begin{equation}\label{prel-3dv-E5}
\begin{aligned}
z_c \begin{bmatrix} u \\ v \\ 1 \end{bmatrix}
&= \begin{bmatrix}
f_x & 0 & c_x \\
0 & f_y & c_y \\
0 & 0 & 1
\end{bmatrix}
\begin{bmatrix}
\bm{R} & \bm{T}
\end{bmatrix}
\begin{bmatrix}
x_e \\
y_e \\
z_e \\
1
\end{bmatrix}, \\
& = \bm{K} \begin{bmatrix}
\bm{R} & \bm{T}
\end{bmatrix}
\begin{bmatrix}
x_e & y_e & z_e & 1
\end{bmatrix}^T .
\end{aligned}
\end{equation}

The matrix $\bm{K} = \begin{bmatrix} f_x & 0 & c_x \\ 0 & f_y & c_y \\ 0 & 0 & 1 \end{bmatrix}$ is known as the camera intrinsics, while the matrix $\begin{bmatrix} \bm{R} & \bm{T} \end{bmatrix}$ represents the camera extrinsics. By leveraging the intrinsics and extrinsics along with the ego-vehicle coordinates of 3D points, it is possible to calculate their projections onto the image plane through the corresponding transformation.

\ifCLASSOPTIONcaptionsoff
  \newpage
\fi

\small
\bibliographystyle{IEEEtran}
\bibliography{IEEEabrv,references}

\newpage

\end{document}